\documentclass[10pt,twocolumn,letterpaper]{article}

\usepackage{iccv}              % To produce the CAMERA-READY version
\usepackage{graphicx}

\newcommand{\rf}[1]{\textcolor{purple}{#1}}
\usepackage[table]{xcolor}
\definecolor{blue}{HTML}{0055cc}
\definecolor{red}{HTML}{cc1100}
\definecolor{orange}{HTML}{cc7700}
\definecolor{green}{HTML}{339955}
\definecolor{Highlight}{HTML}{39a58a}
\definecolor{LightGreen}{HTML}{39a58a}
\usepackage{cuted}
\usepackage{xcolor,soul}
\usepackage{capt-of}
\usepackage{etoc}
\usepackage{algorithm}
\usepackage{algorithmic}
\usepackage{mathtools}
\usepackage{tocloft}
\usepackage{afterpage}
\usepackage[inkscapeformat=png]{svg}
\usepackage[sectionbib]{chapterbib}
\usepackage{caption, subcaption, multirow, overpic, textpos}
\renewcommand{\paragraph}[1]{\vspace{0.75mm}\noindent\textbf{#1}}
\usepackage{array}          % table column width
\usepackage{natbib}
\newcolumntype{x}[1]{>{\centering\arraybackslash}p{#1pt}}
\newcolumntype{y}[1]{>{\raggedright\arraybackslash}p{#1pt}}
\newcolumntype{z}[1]{>{\raggedleft\arraybackslash}p{#1pt}}

\definecolor{iccvblue}{rgb}{0.21,0.49,0.74}
\usepackage{collectbox}
\usepackage[export]{adjustbox}
\usepackage{bbding}         % checkmark
\usepackage{pifont} 
\usepackage{animate}
\usepackage{graphicx}
\usepackage{subcaption}
\usepackage{caption}
\usepackage{comment}
\usepackage{float}
\usepackage{multirow}
\usepackage{subcaption}
\definecolor{my_red}{HTML}{FE4444}
\definecolor{Highlight}{HTML}{39b54a}  % green 
\definecolor{Gray}{gray}{0.95}
\definecolor{LightPurple}{HTML}{845071}
\definecolor{LightYellow}{HTML}{ECE0C2}
\definecolor{MiddleBrown}{HTML}{887F7F}
\definecolor{MiddlePurPle}{HTML}{81506E}
\definecolor{LightGreen}{HTML}{dff0ea}

\newcommand{\tablestyle}[2]{\setlength{\tabcolsep}{#1}\renewcommand{\arraystretch}{#2}\centering\footnotesize}

\usepackage[pagebackref,breaklinks,colorlinks,allcolors=iccvblue]{hyperref}

\def\paperID{8088} % *** Enter the Paper ID here
\def\confName{ICCV}
\def\confYear{2025}

\title{MagicMirror: ID-Preserved Video Generation in Video Diffusion Transformers}

\author{Yuechen Zhang\thanks{Equal contribution} $^{1}$\hspace{0.3cm} Yaoyang Liu\footnotemark[1] $^{2}$\hspace{0.3cm}Bin Xia$^{1}$\hspace{0.3cm}Bohao Peng$^{1}$\hspace{0.3cm}Zexin Yan$^{3}$\hspace{0.3cm}Eric Lo$^{1}$\hspace{0.3cm}Jiaya Jia$^{1,2,4}$\\
$^{1}$CUHK~~~
$^{2}$HKUST~~~
$^{3}$CMU~~~
$^{4}$SmartMore
\small
\vspace{-10pt}
}

\begin{document}
%\maketitle

\twocolumn[{
\maketitle
% \vspace{-35pt}
\centering
%\begin{comment}
\begin{minipage}[b]{0.2485\textwidth}
    \centering
    \animategraphics[width=\linewidth,autoplay,loop,every=2]{6}{figs/gifs/output1_frames_blended/}{0}{47}
\end{minipage}%
\hfill
\begin{minipage}[b]{0.2485\textwidth}
    \centering
    \animategraphics[width=\linewidth,autoplay,loop,every=2]{6}{figs/gifs/output2_frames/}{0}{47}
\end{minipage}%
\hfill
\begin{minipage}[b]{0.2485\textwidth}
    \centering
    \animategraphics[width=\linewidth,autoplay,loop,every=2]{6}{figs/gifs/output7_frames_blended/}{0}{47}
\end{minipage}%
\hfill
\begin{minipage}[b]{0.2485\textwidth}
    \centering
    \animategraphics[width=\linewidth,autoplay,loop,every=2]{6}{figs/gifs/output8_frames/}{0}{47}
\end{minipage}

\begin{minipage}[b]{0.2485\textwidth}
    \centering
    \animategraphics[width=\linewidth,autoplay,loop,every=2]{6}{figs/gifs/output15_frames_blended/}{0}{47}
\end{minipage}%
\hfill
\begin{minipage}[b]{0.2485\textwidth}
    \centering
    \animategraphics[width=\linewidth,autoplay,loop,every=2]{6}{figs/gifs/output9_frames/}{0}{47}
\end{minipage}%
\hfill
\begin{minipage}[b]{0.2485\textwidth}
    \centering
    \animategraphics[width=\linewidth,autoplay,loop,every=2]{6}{figs/gifs/output12_frames_blended/}{0}{47}
\end{minipage}%
\hfill
\begin{minipage}[b]{0.2485\textwidth}
    \centering
    \animategraphics[width=\linewidth,autoplay,loop,every=2]{6}{figs/gifs/output13_frames/}{0}{47}
\end{minipage}\\
%\end{comment}
%\includegraphics[width=0.99\linewidth]{figs/teaser_valid.pdf}
% \vspace{-5pt}
\captionof{figure}{\textbf{MagicMirror generates text-to-video results given the ID reference image.} Complete videos are available in {\scriptsize \textbf{\url{https://julianjuaner.github.io/projects/MagicMirror/}}}.}
\vspace{3mm}
\label{fig:teaser}
}]

% \vspace{-3mm}
\begin{abstract}
We present MagicMirror, a framework for generating identity-preserved videos with cinematic-level quality and dynamic motion. While recent advances in video diffusion models have shown impressive capabilities in text-to-video generation, maintaining consistent identity while producing natural motion remains challenging. Previous methods either require person-specific fine-tuning or struggle to balance identity preservation with motion diversity. Built upon Video Diffusion Transformers, our method introduces three key components: (1) a dual-branch facial feature extractor that captures both identity and structural features, (2) a lightweight cross-modal adapter with Conditioned Adaptive Normalization for efficient identity integration, and (3) a two-stage training strategy combining synthetic identity pairs with video data. Extensive experiments demonstrate that MagicMirror effectively balances identity consistency with natural motion, outperforming existing methods across multiple metrics while requiring minimal parameters added. The code and model will be made publicly available.
\end{abstract}
\vspace{-2mm}
\section{Introduction}
\label{sec:intro}
\vspace{-2mm}

% Human-centered content generation has been a focal point in computer vision research. 
Recent advancements in image generation, particularly through Diffusion Models~\cite{rombach2022high, saharia2022photorealistic, ho2020denoising, betker2023improving}, have propelled personalized content creation to the forefront of computer vision research. While significant progress has been made in preserving personal identity (ID) in image generation~\cite{li2024photomaker,guo2024pulid,wang2024instantid,huang2024consistentid, chen2023photoverse, peng2024portraitbooth}, achieving comparable fidelity in video generation remains challenging.

\begin{figure}[t]
    \centering
    \includegraphics[width=1.0\linewidth]{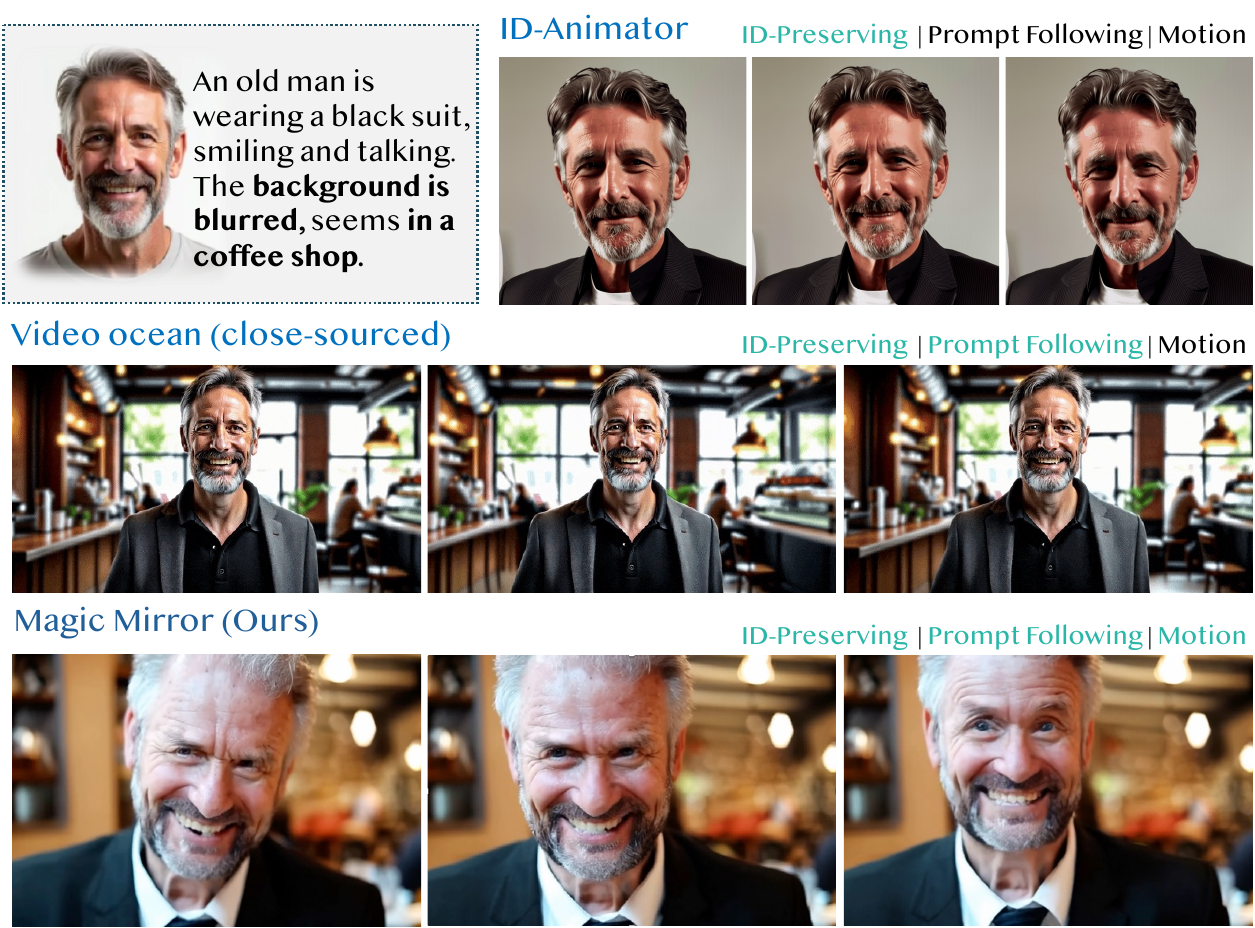}
    \vspace{-4mm}
    \caption{\textbf{MagicMirror generates dynamic facial motion.} ID-Animator~\cite{he2024id} and Video Ocean~\cite{luchen2024ocean} exhibit limited motion range due to a strong identity-preservation constraint. MagicMirror achieves more dynamic facial expressions while maintaining reference identity fidelity.}
    \label{fig:problem}
    \vspace{-5mm}
\end{figure}

Existing ID-preserving video generation methods show promising results but face limitations. Approaches like Magic-Me and ID-Animator~\cite{he2024id,ma2024magic}, utilizing inflated UNets~\cite{guo2023animatediff} for fine-tuning or adapter training, demonstrate some success in maintaining identity across frames. However, they are ultimately restricted by the generation model's inherent capabilities, often failing to produce high-quality videos (see~\cref{fig:problem}). These approaches make a static copy-and-paste instead of generating dynamic facial motions. 
The co-current work ConsisID \cite{yuan2024identity}, implements identity-preserving video generation via a Diffusion Transformer (DiT) framework. While achieving notable performance in video quality, it still exhibits limitations in temporal smoothness and naturalistic facial motions.
Another branch of methods combines image personalization methods with Image-to-Video~(I2V) generation~\cite{yang2024cogvideox,xing2023dynamicrafter,xu2024easyanimate}. While these two-stage solutions preserve ID to some extent, they often struggle with stability in longer sequences and require a separate image generation step.
To address current shortcomings, we present MagicMirror, a single-stage framework designed to generate high-quality videos while maintaining strong ID consistency and dynamic natural facial motions. Our approach leverages native video diffusion models~\cite{yang2024cogvideox} to generate ID-specific videos, aiming to empower individuals as protagonists in their virtual narratives, and bridge the gap between personalized ID generation and high-quality video synthesis.

The generation of high-fidelity identity-preserving videos poses several technical challenges. 
A primary challenge stems from the architectural disparity between image and video generation paradigms. State-of-the-art video generation models, built on full-attention Diffusion Transformer (DiT) architectures~\cite{peebles2023scalable, yang2024cogvideox}, are not directly compatible with conventional cross-attention conditioning methods. To bridge this gap, we introduce a lightweight identity-conditioning adapter integrated into the Video DiT framework. Specifically, we propose a dual-branch facial embedding that simultaneously preserves high-level identity features and reference-specific structural information. Meanwhile, we observed that current video foundation models optimize for text-video alignment, often at the cost of spatial fidelity and generation quality. This trade-off manifests in reduced image quality metrics on benchmarks such as VBench~\cite{huang2024vbench}, particularly introducing the challenge of preserving fine-grained identity features. To address it, we develop a Conditioned Adaptive Normalization~(CAN) that effectively incorporates identity conditions, as a distribution prior, into the pre-trained foundation model. The CAN module, combined with a learnable cross-attention, enables identity conditioning through attention guidance and feature distribution guidance.

Another significant challenge lies in the acquisition of high-quality training data. While paired image data with consistent identity is relatively abundant, obtaining high quality~\cite{yu2023celebv} image-video pairs that maintain high-fidelity identity consistency remains scarce. To address this limitation, we develop a strategic data synthesis pipeline that leverages identity preservation models~\cite{li2024photomaker} to generate paired training data. Our training methodology employs a progressive approach: initially pre-training on image data to learn robust identity representations, followed by video-specific fine-tuning. This two-stage strategy enables effective learning of identity features while ensuring temporal consistency in facial expressions across video sequences.

We evaluate our method on multiple general metrics by constructing a human-centric video generation test set, and comparing it with the aforementioned competitive ID-preserved video generation methods. Extensive experimental and visual evaluations~\cite{van2008visualizing} demonstrate that our approach successfully generates high-quality videos with dynamic content and strong facial consistency, as illustrated in~\cref{fig:teaser}. 
% MagicMirror represents the first approach to achieve customized video generation using Video DiT without requiring person-specific fine-tuning.Our work marks an advancement in personalized video generation, paving the way for enhanced creative expression in the digital domain.
By integrating identity preservation with natural facial motion in Video DiT frameworks without case-specific fine-tuning, MagicMirror advances personalized video generation and enhances creative expression in digital storytelling.

Our main contributions are three-fold: (1) We introduce MagicMirror, a novel fine-tuning free framework with a dual-branch condition extractor for generating ID-preserving videos; (2) We design a lightweight adapter with a conditioned adaptive normalization module, for effective integration of identity features in full-attention Diffusion Transformer architectures; (3) We develop a dataset construction method that combines synthetic data generation with a progressive training strategy to address data scarcity challenges in personalized video generation.

\section{Related Works}
\label{sec:related_work}
\paragraph{Diffusion Models.}
Since the introduction of DDPM~\cite{ho2020denoising}, diffusion models have demonstrated remarkable capabilities across diverse domains, spanning NLP~\cite{li2022diffusion, gong2022diffuseq}, medical imaging~\cite{chung2022score, chung2022come}, and molecular modeling~\cite{corso2022diffdock, jing2022torsional}. In computer vision, following initial success in image generation~\cite{dhariwal2021diffusion, kingma2021variational}, Latent Diffusion Models (LDM)~\cite{rombach2022high} significantly reduced computational requirements while maintaining generation quality. Subsequent developments in conditional architectures~\cite{ruiz2023dreambooth, nichol2021glide} enabled fine-grained concept customization over the generation process.

\paragraph{Video Generation via Diffusion Models.}
Following the emergence of diffusion models, their superior controllability and diversity in image generation~\cite{xing2023survey} have led to their prominence over traditional approaches based on GANs~\cite{goodfellow2020generative, karras2020analyzing, karras2019style} and auto-regressive Transformers~\cite{esser2021taming, ramesh2021zero, yu2022scaling}. The Video Diffusion Model (VDM)~\cite{ho2022video} pioneered video generation using diffusion models by extending the traditional U-Net~\cite{ronneberger2015u} architecture to process temporal information. Subsequently, LVDM~\cite{he2022latent} demonstrated the effectiveness of latent space operations, while AnimateDiff~\cite{guo2023animatediff} adapted text-to-image models for personalized video synthesis. A significant advancement came with the Diffusion Transformer (DiT)~\cite{peebles2023scalable}, which successfully merged Transformer architectures~\cite{vaswani2017attention, dosovitskiy2020image} with diffusion models. Building on this foundation, Latte~\cite{ma2024latte} emerged as the first open-source text-to-video model based on DiT. Following the breakthrough of SORA~\cite{2024sora}, several open-source initiatives including Open-Sora-Plan~\cite{lin2024opensoraplanopensourcelarge}, Open-Sora~\cite{opensora}, and CogVideoX~\cite{yang2024cogvideox} have advanced video generation through DiT architectures. While current research predominantly focuses on image-to-video translation~\cite{xing2023dynamicrafter, xu2024easyanimate, zeng2023makepixelsdancehighdynamic} efficiency, and motion control~\cite{wang2024motionctrl,yu2024viewcrafter,xing2025motioncanvas,zhang2025training}, the critical challenge of ID-preserving video generation remains relatively unexplored.

\paragraph{ID-Preserving Generation.}
ID-preserving generation, or identity customization, aims to maintain specific identity characteristics in generated images or videos. Initially developed in the GAN era~\cite{goodfellow2020generative} with significant advances in face generation~\cite{richardson2021encoding, wang2021towards, karras2019style}, this field has evolved substantially with diffusion models, demonstrating enhanced capabilities in novel image synthesis~\cite{ruiz2023dreambooth, gal2022image}. Current approaches to ID-preserving image generation fall into two main categories:

\noindent\textit{Tuning-based Models}: These approaches fine-tune models using one or more reference images to generate identity-consistent outputs. Notable examples include Textual Inversion~\cite{gal2022image} and Dreambooth~\cite{ruiz2023dreambooth}.

\noindent\textit{Tuning-free Models}: Addressing the computational overhead problem, these models maintain high ID fidelity through additional conditioning and trainable parameters. Starting with IP-adapter~\cite{ye2023ip}, various methods such as InstantID, PhotoMaker~\cite{wang2024instantid, li2024photomaker, chen2023photoverse, peng2024portraitbooth, huang2024consistentid, guo2024pulid} have emerged to enable efficient and high quality personalized generation.

ID-preserving video generation introduces additional complexities, particularly in synthesizing realistic facial movements from static references while maintaining identity consistency. Current approaches include Magic-Me~\cite{ma2024magic}, a tuning-based method requiring per-identity optimization, and ID-Animator~\cite{he2024id}, a tuning-free approach utilizing face adapters and decoupled Cross-Attention~\cite{ye2023ip}. However, these methods face challenges in maintaining dynamic expressions while preserving identity, and are constrained by base model limitations in video quality, duration, and prompt adherence. The integration of Diffusion Transformers presents promising opportunities for advancing ID-preserving video generation. 
ConsisID~\cite{yuan2024identity} is the co-current work with the DiT architecture, but it faces the problem of video smoothness and face dynamics.

\begin{figure*}
    \centering
    \includegraphics[width=1.0\linewidth]{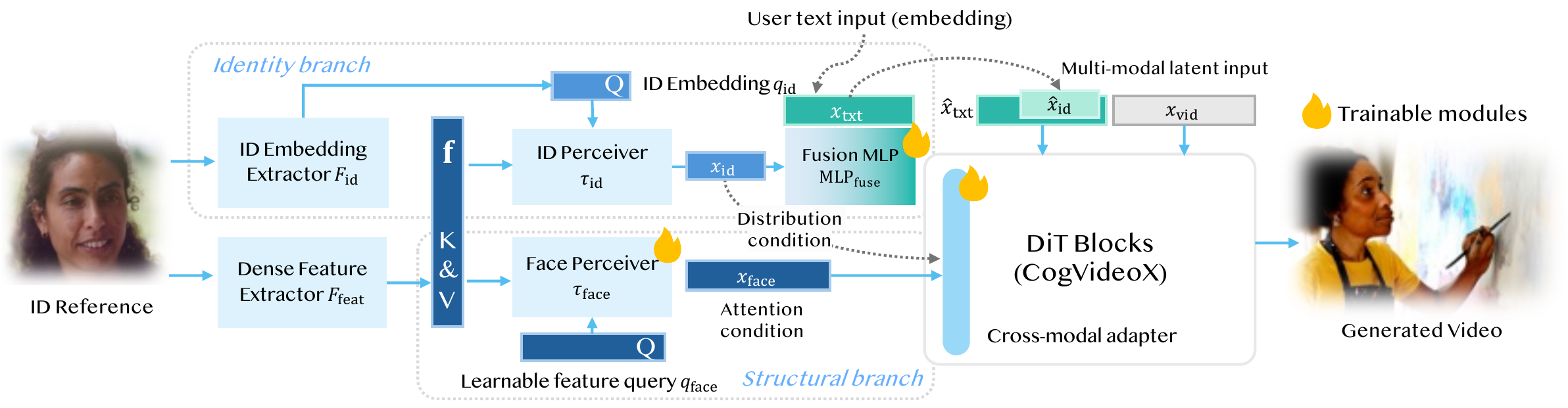}
    \vspace{-5mm}
    \caption{\textbf{Overview of MagicMirror.} The framework employs a dual-branch feature extraction system with ID and face perceivers, followed by a cross-modal adapter~(illustrated in \cref{fig:adapter}) for DiT-based video generation. By optimizing trainable modules marked by the flame, our method efficiently integrates facial features for controlled video synthesis while maintaining model efficiency.}
    \label{fig:pipeline}
    \vspace{-5mm}
\end{figure*}

\vspace{-3mm}
\section{MagicMirror}
\vspace{-1mm}
\label{sec:method}

\setlength{\abovedisplayskip}{3pt}
\setlength{\belowdisplayskip}{3pt}

An overview of MagicMirror is illustrated in \cref{fig:pipeline}. This dual-branch framework~(\cref{sec:dual}) extracts facial identity features from one or more reference images $r$. They are subsequently processed through a DiT backbone augmented with a lightweight cross-modal adapter, incorporating Conditioned Adaptive Normalization (\cref{sec:CAN}). This architecture enables MagicMirror to synthesize identity-preserved text-to-video outputs. The following sections elaborate on the preliminaries of diffusion models (\cref{sec:preliminary}) and each component of our method.

\vspace{-1mm}
\subsection{Preliminaries}
\vspace{-1mm}
\label{sec:preliminary}
Latent Diffusion Models (LDMs) generate data by iteratively reversing a noise corruption process, converting random noise into structured samples. At time step $t \in \{0, \ldots, T\}$, the model predicts latent state $x_t$ conditioned on $x_{t+1}$:
\begin{equation}
    p_{\theta}(x_t|x_{t+1}) = \mathcal{N}(x_t;\widetilde{\mu}_t,\widetilde{\beta}_t \mathit{I}),
\end{equation}
where $\theta$ represents the model parameters, $\widetilde{\mu}_t$ denotes the predicted mean, and $\widetilde{\beta}_t$ is the variance schedule.

The training objective typically employs a mean squared error loss $\mathcal{L}_\text{noise}$ on the noise prediction $\epsilon_\theta(x_t,t,c_\text{txt})$:
\begin{equation}
\mathcal{L}_\text{noise} =  \mathbb{E}_{t,c_\text{txt},\epsilon \sim \mathcal{N}(0,1)}\Big[\Vert 
    \epsilon - \epsilon_\theta(x_t,t,c_\text{txt})\Vert^2\Big]\text{,}
    \label{eq:diff_loss}
\end{equation}
where $c_\text{txt}$ denotes the text condition.

Recent studies on controllable generation~\cite{zhang2023adding, peng2024controlnext, ye2023ip, wang2024instantid} extend this framework by incorporating additional control signals, such as image condition $c_{\text{img}}$. This is achieved through a feature extractor $\tau_{\text{img}}$ that processes a reference image $r$: $c_{\text{img}} = \tau_{\text{img}}(r)$. Consequently, the noise prediction function in~\cref{eq:diff_loss} becomes $\epsilon_\theta(x_t,t,c_\text{txt}, c_{\text{img}})$. In this paper, we denote the additional facial condition by $c_{\text{face}}$.

\vspace{-1mm}
\subsection{Dual-Branch Facial Feature Extraction}
\vspace{-1mm}
\label{sec:dual}
The facial feature extraction module in MagicMirror, depicted in the left part of~\cref{fig:pipeline}, is designed to extract high-level identity information and detailed structural features. Given an identity reference image $r \in \mathbb{R}^{h \times w \times 3}$, our model extracts facial condition embeddings $c_{\text{face}} = \{x_{\text{face}}, \hat{x}_{\text{id}}\}$ using a dual-branch perceiver architecture, where each branch specializes in a distinct aspect of facial representation.

To obtain these embeddings, we first extract dense feature maps $\mathbf{f}= F_{\text{feat}}(r)$ from the input image, where $F_{\text{feat}}$ is a pre-trained CLIP ViT encoder~\cite{radford2021learning} that captures rich facial semantics. The \textbf{identity branch} employs an identity perceiver $\tau_{\text{id}}$ to extract high-level identity features:
\begin{equation}
    x_{\text{id}} = \tau_{\text{id}}(q_{\text{id}}, \mathbf{f})\text{,}
    \label{eq:x_id}
\end{equation}
where $q_\text{id} = F_\text{id}(r)$ represents high-level identity-aware features extracted by an ArcFace encoder $F_\text{id}$~\cite{deng2019arcface, li2024photomaker}.

The \textbf{structural branch} utilizes a facial structure perceiver $\tau_{\text{face}}$ that focuses on fine-grained facial details, leveraging a learnable query embedding $q_{\text{face}}$ for facial structure extraction:
\begin{equation}
    x_{\text{face}} = \tau_{\text{face}}(q_{\text{face}}, \mathbf{f})\text{.}
    \label{eq:x_face}
\end{equation}
Both perceivers $\tau_{\text{face}}$ and $\tau_{\text{id}}$ implement the standard Q-Former architecture~\cite{li2023blip} with distinct query conditions in~\cref{eq:x_id,eq:x_face}.
The identity branch and structural branch serve complementary roles in ensuring high-quality ID-preserving video generation. The identity branch maintains consistent identity features across frames, guaranteeing that the generated video preserves the subject's identity throughout the sequence. Conversely, the structural branch captures fine-grained facial details and facilitates dynamic natural facial motions, effectively preventing structural collapse during movement.

To control these aspects effectively, we employ distinct feature injection mechanisms. Identity information is incorporated into the text representation, building upon recent advancements in personalized text-to-image generation~\cite{ruiz2023dreambooth, li2024photomaker}. Specifically, identity embeddings are projected into the text embedding space via a fusion MLP $\text{MLP}_{\text{fuse}}$, $\hat{x}_{\text{id}} = \text{MLP}_{\text{fuse}}(x_{\text{id}}, \mathbf{m} x_{\text{txt}})$,
where $x_{\text{txt}}$ denotes the input textual prompt embedding, $\mathbf{m}$ is a token-level binary mask that selectively applies fusion at identity-relevant tokens (e.g., ``man", ``woman") within $x_{\text{txt}}$. This process results in a fused id representation $\hat{x}_{\text{id}}$. The final adapted text embedding for DiT is computed as a masked replacement:
\begin{equation}
    \hat{x}_{\text{txt}} = \mathbf{m} \hat{x}_{\text{id}} + (1 - \mathbf{m}) x_{\text{txt}},
\end{equation}
Meanwhile, the structural embedding $x_{\text{face}}$ is utilized as direct conditioning signals to guide the generative process. 
%, ensuring realistic and coherent facial motion throughout the video.

\subsection{Conditioned Adaptive Normalization}
\label{sec:CAN}

Having obtained the decoupled ID-aware conditions \(c_\text{face}\), we address the challenge of efficiently integrating these conditions into the video diffusion transformer. Traditional Latent Diffusion Models, as exemplified by Stable Diffusion~\cite{rombach2022high}, utilize isolated cross-attention mechanisms for condition injection, which allow for straightforward adaptation to new conditions via decoupled cross-attention~\cite{ye2023ip,guo2024pulid}. 
%This approach is based on uniform conditional inputs—specifically, the text condition \(c_\text{txt}\)—across all cross-attention layers. 
However, our framework is based on mm-DiTs~\cite{yang2024cogvideox}, which implements a cross-modal full-attention paradigm coupled with layer-wise distribution modulation experts. This architectural choice introduces additional complexity in adapting to new conditions beyond simple cross-attention augmentation. Under this constraint, we find that \textit{\textbf{modulation layers, despite requiring extremely few parameters, play a crucial role in learning the data distribution.}} This point is discussed experimentally in the Appendix \textcolor{iccvblue}{B.2}. 

\begin{figure}[t]
    \centering
    \includegraphics[width=1.0\linewidth]{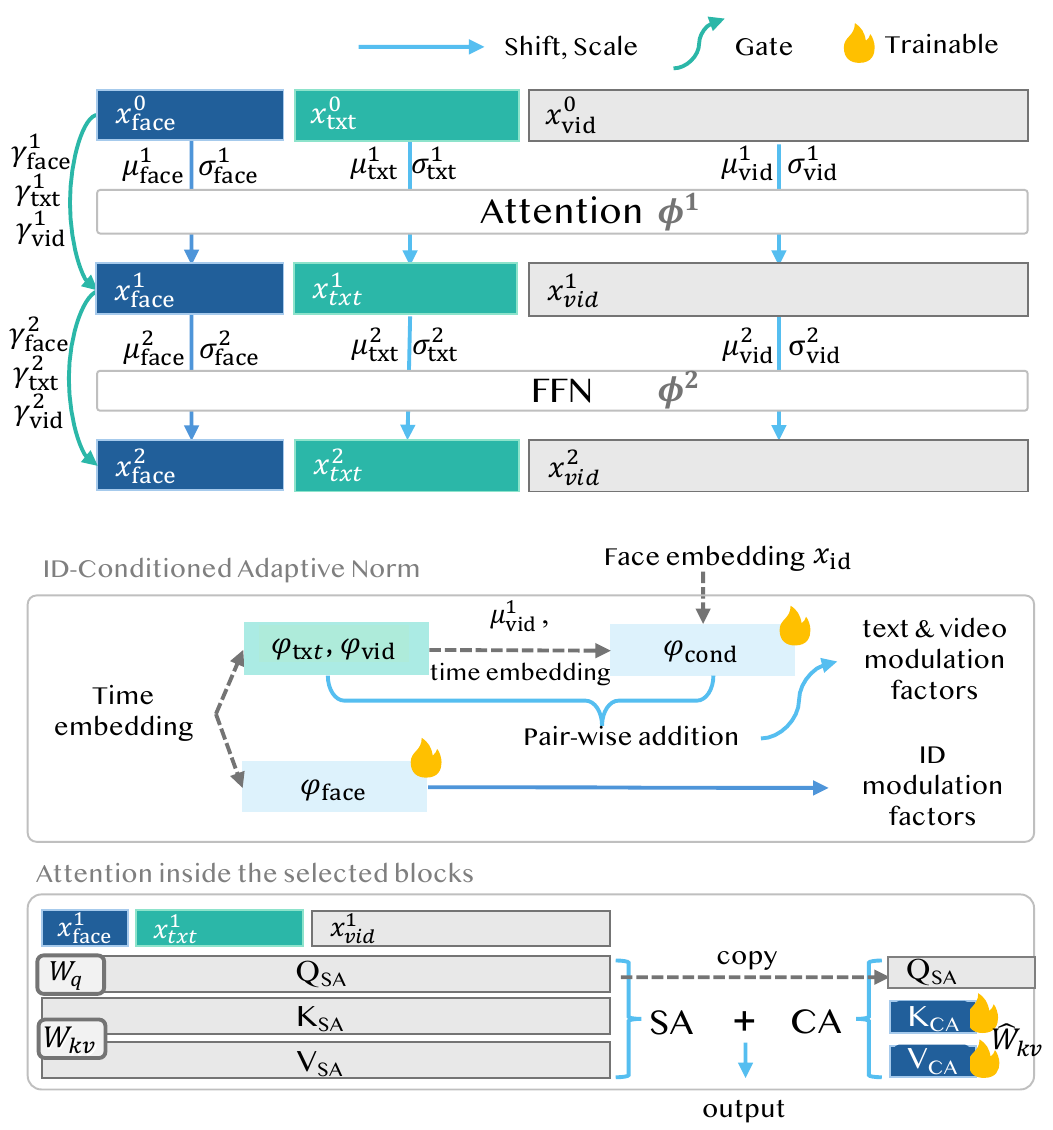}
    \vspace{-4mm}
    \caption{\textbf{Cross-modal adapter} in DiT blocks. \textit{Top:}  Cross-modal modulation in mmDiTs. \textit{Bottom:} The Conditioned Adaptive Normalization (CAN) for modal-specific feature modulation and decoupled attention integration.}
    \label{fig:adapter}
    \vspace{-4mm}
\end{figure}

Leveraging mm-DiT's layer-wise modulation~\cite{yang2024cogvideox}, we propose a lightweight adapter that incorporates additional facial conditions. As illustrated in~\cref{fig:adapter}, facial embedding $x_\text{face}$ in~\cref{eq:x_face} is concatenated with text and video features ($x_\text{txt}$ and $x_\text{vid}$) to feed into the full self-attention. CogVideoX employs modal-specific modulation, where factors $m_\text{vid}$ and $m_\text{txt}$ are applied to their respective modalities through adaptive normalization, where modulation factors are extracted from MLP extractors $\varphi_{\{\text{txt, vid}\}}$. To accommodate the facial modality, we introduce a dedicated adaptive normalization module, normalizing facial features preceding the self-attention and feed-forward network (FFN). The corresponding set of modulation factors for the facial modality $m_\text{face}$ is computed by a MLP  $\varphi_\text{face}$:
\begin{equation}
m_\text{face} = \{\mu^1_\text{face}, \sigma^1_\text{face}, \gamma^1_\text{face}, \mu^2_\text{face}, \sigma^2_\text{face}, \gamma^2_\text{face}\} = \varphi_\text{face}(\mathbf{t}, l),
\label{eq:ln}
\end{equation}
where $\mathbf{t}$ denotes the time embedding and $l$ represents the layer index. Here, \(\mu\) is the shift parameter to adjust the feature mean, \(\sigma\) is the scale parameter to modulating the feature amplitude, and \(\gamma\) is the gating parameter to control the influence of the modulation.
Within each block, let \(\phi^n\) denote the in-block operations—where \(\phi^1\) represents the self-attention operation and \(\phi^2\) represents the FFN. The feature transformation after operation \(n\) is computed as:
$
\bar{x}^{n} = x^{n-1} * (1 + \sigma^n) + \mu^n $, then $
x^{n} = \bar{x}^{n} + \gamma^n \phi^{n}(\bar{x}^n)
$, with modality-specific subscripts omitted for brevity.

Furthermore, we explore whether layer-wise distribution conditioned by ID features would make ID injection more effective. To enhance the distribution learning capability of text and video latents from specific reference IDs, we introduce Conditioned Adaptive Normalization (CAN), inspired by class-conditioned DiT~\cite{peebles2023scalable} and StyleGAN's~\cite{karras2019style} approach to control with conditions. Based on $\varphi_{\{\text{txt, vid}\}}$, CAN predicts distribution shifts for video and text modalities with a trainable MLP $\varphi_\text{cond}$:
\begin{equation}
\hat{m}_\text{vid}, \hat{m}_\text{txt} = \varphi_\text{cond}(\mathbf{t}, l, \mu^1_\text{vid}, x_\text{id}).
\label{eq:ln_add}
\end{equation}
Here, $\mu^1_\text{vid}$ acts as a distribution identifier for a better initialization of the CAN module, and $x_\text{id}$ from~\cref{eq:x_id} represents the identity distribution prior. The final modulation factors are computed via residual addition: $m_\text{vid} = \hat{m}_\text{vid} + \varphi_\text{vid}(\mathbf{t}, l), m_\text{txt} = \hat{m}_\text{txt} + \varphi_\text{txt}(\mathbf{t}, l)$. 
% We found this conditional shift prediction $\varphi_\text{cond}$ is well-suited to an MLP implementation.

Complementing the conditioned normalization, we augment the joint full self-attention $T_\text{SA}$ with a cross-attention mechanism $T_\text{CA}$~\cite{ye2023ip, he2024id} to further enhance the aggregation of ID modality features. The final attention output is computed as: 
\begin{equation}
\footnotesize
{
x_\text{out} = T_\text{SA}(W_{q}(x_\text{full}), W_{kv}(x_\text{face})) + T_\text{CA}(W_{q}(x_\text{full}), \hat{W}_{kv}(x_\text{face}))\text{,}
}
\label{eq:attention}
\end{equation}
where $x_\text{full}$ denotes the complete aggregated feature representation produced by concatenating or integrating text, video, and facial features from preceding layers. $T_\text{SA}$ and $T_\text{CA}$ utilize the same query projection $W_{q}(x_\text{full})$, while the key-value projections $\hat{W}_{kv}$ in cross-attention are re-initialized and trainable.

\subsection{Data and Training}
\begin{figure}[t]
    \centering
    \includegraphics[width=1.0\linewidth]{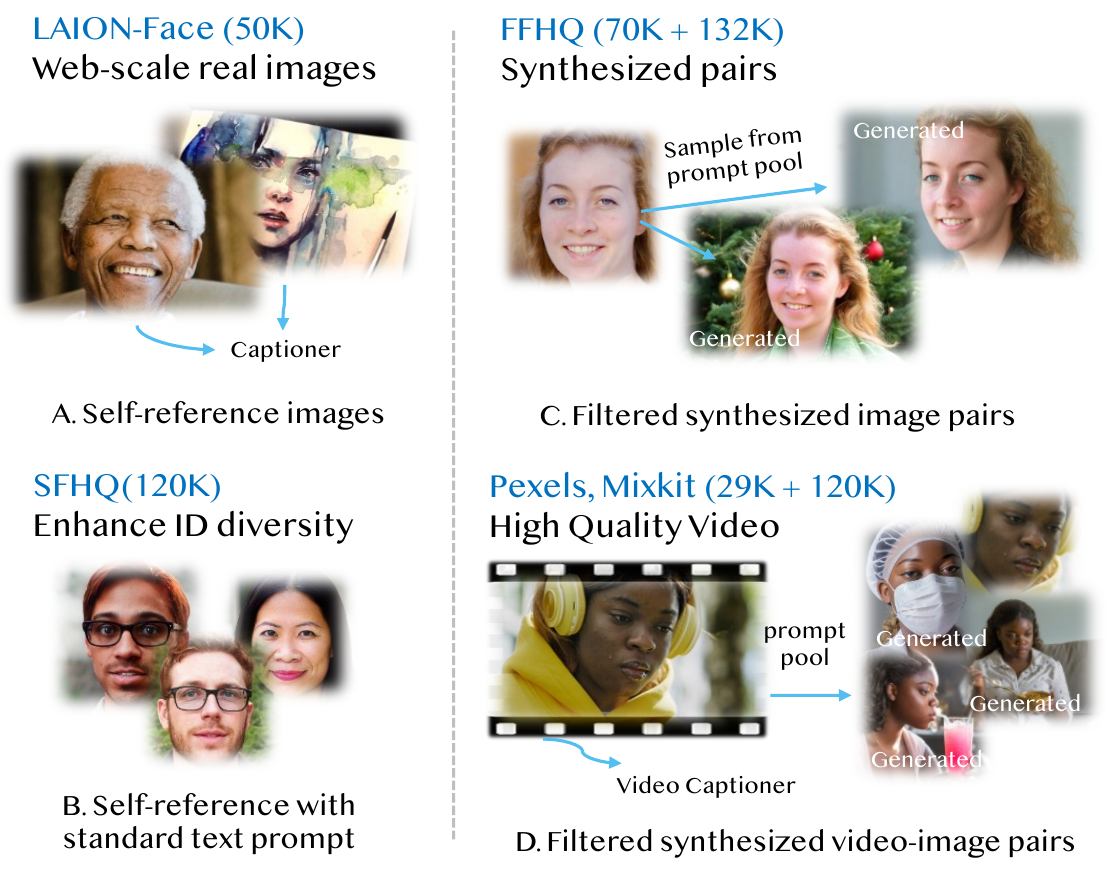}
    \vspace{-3mm}
    \caption{\textbf{Overview of our training datasets.} The pipeline includes image pre-training data (A-C) and video post-training data (D). We utilize both self-reference data (A, B) and filtered synthesized pairs with the same identity~(C, D). Numbers of (images + synthesized images) are reported.}
    \label{fig:data}
    \vspace{-4mm}
\end{figure}

Training a zero-shot customization adapter presents unique data challenges compared to fine-tuning approaches, like Magic-Me~\cite{ma2024magic}. Our model's full-attention architecture, which integrates spatial and temporal components inseparably, necessitates a two-stage training strategy. As shown in~\cref{fig:data}, we begin by training on diverse, high-quality datasets to develop robust identity preservation capabilities.

Our progressive training pipeline leverages diverse datasets to enhance model performance, particularly in identity preservation. For image pre-training, we first utilize the LAION-Face~\cite{schuhmann2021laion} dataset, which contains web-scale real images and provides a rich source for generating self-reference images. To further increase identity diversity, we utilize the SFHQ~\cite{david_beniaguev_2022_SFHQ} dataset, which applies self-reference techniques with standard text prompts. To prevent overfitting and promote the generation of diverse face-head motion, we use the FFHQ~\cite{karras2019style} dataset as a base. From this, we random sample text prompts from a prompt pool of human image captions, and synthesize ID-conditioned image pairs using PhotoMaker-V2~\cite{li2024photomaker}, ensuring both identity similarity and facial motion diversity through careful filtering.

For video post-training, we leverage the high-quality Pexels and Mixkit datasets~\cite{pexels,mixkit}, along with a small collection of self-collected videos from the web. Similarly, synthesized image data corresponding to each face reference of keyframes are generated as references. The combined dataset offers rich visual content for training the model across images and videos.

The objective function combines identity-aware and general denoising loss:
$
\mathcal{L} = \mathcal{L}_{\text{noise}} + \lambda \left(1 - \cos(q_{\text{face}}, D(x_0)) \right)\text{,}
$
where $D(\cdot)$ represents the latent decoder for the denoised latent $x_0$, and $\lambda$ is the balance factor. Following PhotoMaker~\cite{li2024photomaker}, we compute the denoising loss specifically within the face area for 50\% of random training samples.

\begin{table*}[ht!]
    \centering
    \scriptsize
    \tablestyle{2.5pt}{1.1}
    \begin{tabular*}{\linewidth}
    {y{75}|x{35}x{40}x{40}x{45}|x{45}x{35}x{45}x{45}|x{40}}
    \toprule
\multirow{2}{*}{Models} & Dynamic Degree\rf{$^\uparrow$} &  Text Alignment\rf{$^\uparrow$} & Inception Score\rf{$^\uparrow$} & Motion Smoothness\rf{$^\downarrow$} & Average ID Similarity\rf{$^\uparrow$} & Similarity Decay\rf{$^\downarrow$} & Face Motion FM$_\text{ref}$\rf{$^\uparrow$} &Face Motion FM$_\text{inter}$\rf{$^\uparrow$} & Overall Preference\rf{$^\uparrow$}\\
\midrule
DynamiCrafter~\cite{xing2023dynamicrafter}&    0.455 &   0.168 & 8.20 & 0.507 & 0.896 & \textbf{0.002} & 0.237 & 0.287 & 5.402
\\ 
EasyAnimate-I2V~\cite{xu2024easyanimate}&   0.155 &   0.177 & 9.55 & \textbf{0.482} &0.903 & 0.022 & 0.262 & 0.278 & 5.935
\\ 
CogVideoX-I2V~\cite{ yang2024cogvideox}&  \underline{0.660} &   0.213 & 9.85 & 0.497 & 0.901 & 0.029 & 0.413 & 0.532 &  \underline{6.985} 
\\ 
ID-Animator~\cite{he2024id}&    0.140 &  0.211 & 7.57 & 0.515 & \textbf{0.923} & 0.005 & \underline{0.652} & 0.181 & 5.693 
\\ 
ConsisID~\cite{yuan2024identity}& 0.615 &  \underline{0.236} & \textbf{11.09} & 0.513 & 0.913 & \textbf{0.002} & \underline{0.652} & \underline{0.601} & 6.640
\\ 
\cellcolor{Gray}
\textbf{MagicMirror}& \cellcolor{Gray}\textbf{0.705} & \cellcolor{Gray}\textbf{0.240} & \cellcolor{Gray}\underline{10.59} & \cellcolor{Gray}\underline{0.484} 
& \cellcolor{Gray}\underline{0.922} & \cellcolor{Gray}\textbf{0.002} & \cellcolor{Gray}\textbf{0.730} & \cellcolor{Gray}\textbf{0.610} & \cellcolor{Gray}\textbf{7.315} \\

    \bottomrule
    \end{tabular*}
    \vspace{-2mm}
    \caption{\textbf{Quantitative comparisons.} We report results with Image-to-Video and ID-preserved models. ID similarities are evaluated on the corresponding face-enhanced prompts to avoid face missing caused by complex prompts. Arrows indicate the direction of improved performance for each metric. We highlight the \textbf{best} and the \underline{second best} results for each metric.
    }
    \vspace{-5mm}
    \label{tab:quantitative_result}
\end{table*}
\vspace{-1mm}
\section{Experiments}
\vspace{-1mm}
\label{sec:experiments}

\subsection{Implementation Details}
\vspace{-1mm}
\paragraph{Dataset preparation.} As illustrated in~\cref{fig:data}, our training pipeline leverages both self-referenced and synthetically paired image data~\cite{karras2019style, david_beniaguev_2022_SFHQ, schuhmann2021laion} for identity-preserving alignment in the initial training phase. For synthetic data pairs (denoted as C and D in~\cref{fig:data}), we employ ArcFace~\cite{deng2019arcface} for facial recognition and detection to extract key attributes including age, bounding box coordinates, gender, and facial embeddings. Reference frames are then generated using PhotoMakerV2~\cite{li2024photomaker}. We implement a quality control process by filtering image pairs $\{a, b\}$ based on their facial embedding cosine similarity, retaining pairs where $cos(q^a_\text{face}, q^b_\text{face}) > 0.65$, $q_{\text{face}}$ means the facial embedding. %This results in a curated dataset of 130K images comprising 49K unique identities. 
For text conditioning, we utilize MiniGemini-8B~\cite{li2024mini} to caption all video data, to form a diverse prompt pool containing 29K prompts, while CogVLM~\cite{wang2023cogvlm} provides video descriptions in the second training stage. Detailed data collection procedures are provided in Appendix \textcolor{iccvblue}{A.1}.
%~\cref{sec:data_collection}.

\paragraph{Training Details.}
Our MagicMirror framework extends CogVideoX-5B~\cite{yang2024cogvideox} by integrating facial-specific modal adapters into alternating DiT layers (i.e., adapters in all layers with even index $l$). We adopt the feature extractor $F_\text{feat}$ and ID perceiver $\tau_\text{id}$ from a pre-trained PhotoMakerV2~\cite{li2024photomaker}. 
%The training procedure comprises two phases. 
In the image pre-train stage, we optimize the adapter components for 30K iterations using a global batch size of 64. Subsequently, we perform video fine-tuning for 5K iterations with a batch size of 8 to enhance temporal consistency in video generation. Both phases employ a decayed learning rate starting from $10^{-5}$. All experiments were conducted on a single compute node with 8 NVIDIA A800 GPUs.

\begin{figure*}[t]
    \centering
    \includegraphics[width=0.995\linewidth]{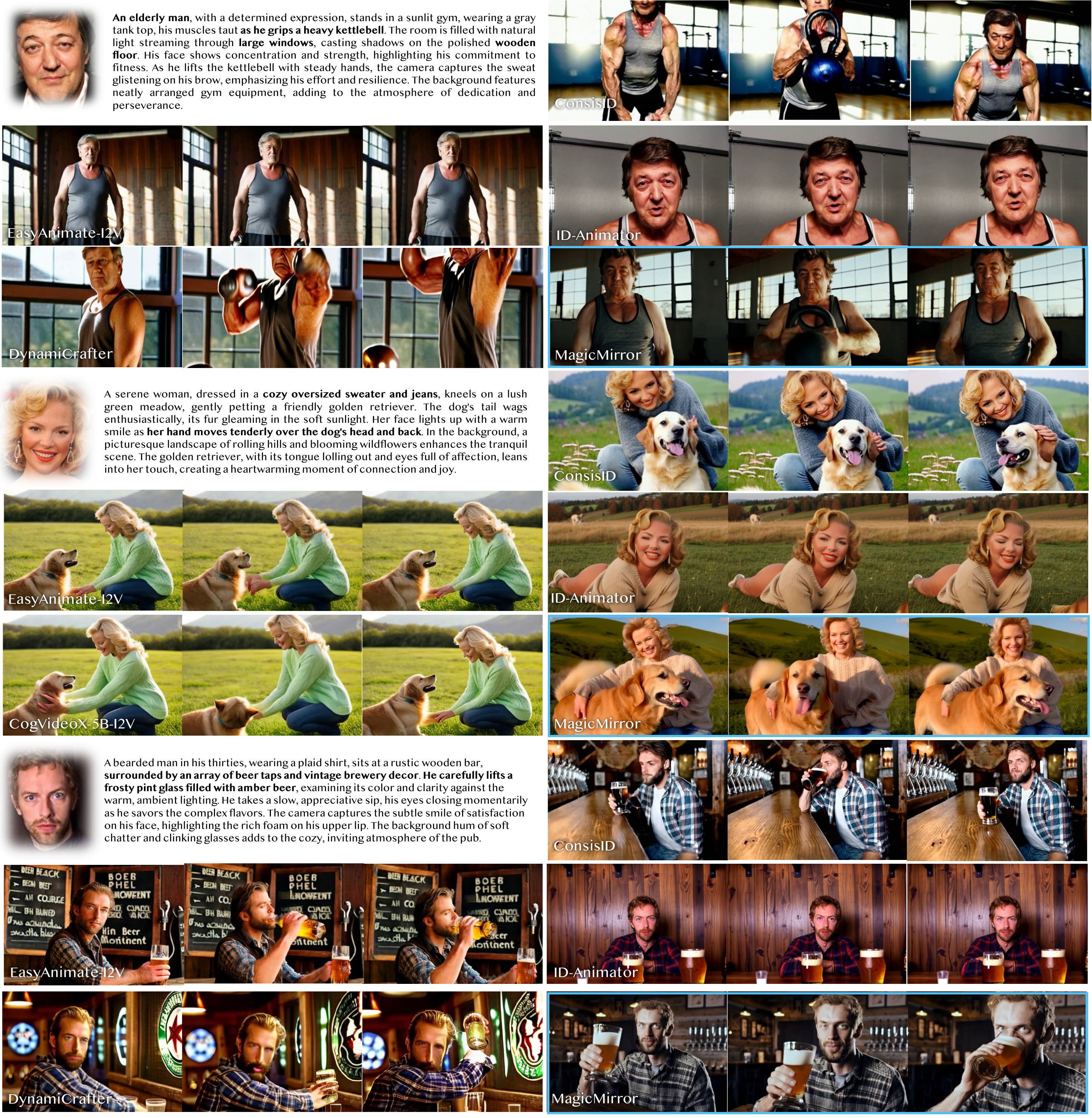}
    \vspace{-3mm}
    \caption{\textbf{Qualitative comparisons.} Captions and reference identity images are presented in the top-left corner for each case. 
    %For video results, please refer to the Appendix.
    }
    \label{fig:compare}
    \vspace{-3mm}
\end{figure*}

\paragraph{Evaluation and Comparisons.}
We evaluate our approach against the state-of-the-art ID-consistent video generation model ID-Animator~\cite{he2024id}, ConsisID~\cite{yuan2024identity} and leading Image-to-Video (I2V) frameworks, including DynamiCrafter~\cite{xing2023dynamicrafter}, CogVideoX-I2V~\cite{yang2024cogvideox}, and EasyAnimate~\cite{xu2024easyanimate}. Our evaluation leverages standardized VBench~\cite{huang2024vbench}, for video generation assessment that measures motion quality and text-motion alignment. For identity preservation, we utilize facial recognition embedding similarity~\cite{geitgey2017facerecognition} and facial motion metrics. Our evaluation dataset consists of 40 single-character prompts from VBench, ensuring demographic diversity, and 40 action-specific prompts for motion assessment. Identity references are sampled from 50 face identities from PubFig~\cite{kumar2009attribute}, generating four personalized videos per identity across varied prompts.

\vspace{-1mm}
\subsection{Quantitative Evaluation}
\vspace{-1mm}
The quantitative results are summarized in~\cref{tab:quantitative_result}. We evaluate generated videos using VBench's and EvalCrafter's general metrics~\cite{huang2024vbench, liu2024evalcrafter}, including dynamic degree, text-prompt consistency, and Inception Score~\cite{salimans2016improved} for video quality assessment. We also evaluate on smoothness using cross-frame optical flow consistency. For identity preservation, we introduce Average Similarity, measuring the distance between generated faces and the average similarity of a group of reference images with the same identity. This prevents models from achieving artificially high scores through naive copy-paste behavior, as illustrated in~\cref{fig:problem}. Face motion is quantified using two metrics: FM$_\text{ref}$ (relative distance to the reference face) and FM$_\text{inter}$ (inter-frame distance), computed using RetinaFace~\cite{deng2020retinaface} landmarks after position alignment, and the L2 distance between the normalized coordinates is reported as the metric.

Our method achieves superior facial similarity compared to I2V approaches while maintaining competitive performance to ID-Animator and ConsisID. We demonstrate strong text alignment, video quality, and dynamic performance, attributed to our decoupled facial feature extraction and cross-modal adapter with CAN.

Besides, we analyze facial similarity drop across uniformly sampled frames from each video to assess temporal identity consistency, reporting as the similarity decay term in~\cref{tab:quantitative_result}. Standard I2V models (CogVideoX-I2V~\cite{yang2024cogvideox}, EasyAnimate~\cite{xu2024easyanimate}) exhibit significant temporal decay in identity preservation. Although DynamiCrafter~\cite{xing2023dynamicrafter} shows better stability due to its random reference strategy, it compromises fidelity. Both ConsisID~\cite{yuan2024identity} and MagicMirror maintain consistent identity preservation throughout the video duration.

%, while our method showed smaller discrepancies in RS and AS metrics, indicating that our method captures more facial information, resulting in better generalization capability.
\begin{table}
\tablestyle{2.7pt}{1.1}
	\centering
        \scriptsize
	{
		\begin{tabular}{y{75}|x{30}x{38}x{30}x{38}}
			\toprule
 			\multirow{2}{*}{Models} & Visual Quality\rf{$^\uparrow$} &  Text Alignment\rf{$^\uparrow$} & Dynamic Degree\rf{$^\uparrow$} & ID Similarity\rf{$^\uparrow$} \\
    %[6.035532994923858, 7.289340101522843, 4.852791878172589, 3.431472081218274]
			\midrule
			DynamiCrafter~\cite{xing2023dynamicrafter} & 6.03 & 7.29 & 4.85  & 5.87\\
    %[6.629441624365482, 8.208121827411167, 5.573604060913706, 3.3299492385786804]
                EasyAnimate-I2V~\cite{xu2024easyanimate}  & 6.62 & 8.21 & 5.57 & 6.01 \\
    %[6.862944162436548, 8.314720812182742, 6.548223350253807, 3.8883248730964466]
                CogVideoX-I2V~\cite{yang2024cogvideox} & \underline{6.86} & \underline{8.31} & \underline{6.55} & 6.22\\
    %[5.6395939086294415, 6.365482233502538, 4.060913705583756, 5.137055837563452]
                ID-Animator~\cite{he2024id} & 5.63 & 6.37 & 4.06 & \textbf{6.70} \\
                ConsisID~\cite{yuan2024identity} & 6.43 & 8.35 & 6.23 & 5.55 \\
    %[6.969849246231155, 8.884422110552764, 7.015075376884422, 6.386363636363637]
                \cellcolor{Gray}\textbf{MagicMirror} & \cellcolor{Gray}\textbf{6.97} & \cellcolor{Gray}\textbf{8.88}& \cellcolor{Gray}\textbf{7.02} & \cellcolor{Gray}\underline{6.39}\\
			\bottomrule
		\end{tabular}}
        \vspace{-2mm}
        \caption{\textbf{User study results.} We highlight the \textbf{best} and the \underline{second best} results for each metric.}
        \vspace{-3mm}
        \label{tab:userstudy}
\end{table}

\begin{figure*}[t]
    \centering
    \includegraphics[width=0.99\linewidth]{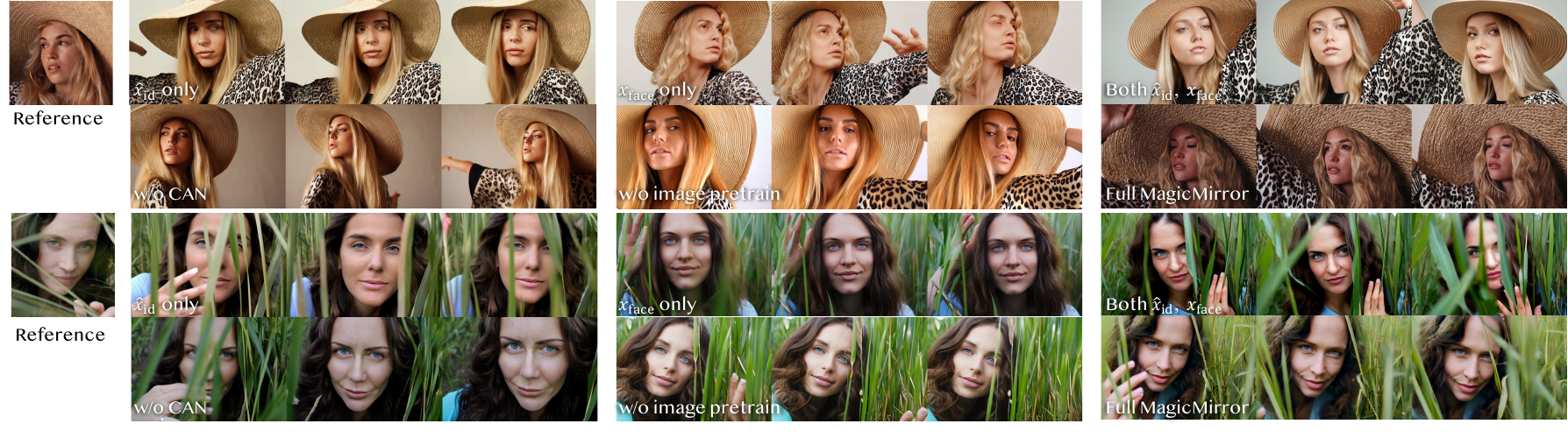}
    \vspace{-4mm}
    \caption{\textbf{Examples for ablation studies} on modules and training strategies.}
    \label{fig:ablation}
    \vspace{-5mm}
\end{figure*}

\vspace{-1mm}
\subsection{Qualitative Evaluation}
\vspace{-1mm}
Beyond the examples shown in~\cref{fig:teaser}, we present comparative results in~\cref{fig:compare}. Our method maintains high text coherence, motion dynamics, and video quality compared to conventional CogVideoX inference. When compared to existing image-to-video approaches~\cite{li2024photomaker, xu2024easyanimate, xing2023dynamicrafter, yang2024cogvideox}, MagicMirror demonstrates superior identity consistency across frames while preserving natural motion. Our method also achieves enhanced dynamic range and text alignment compared to ID-Animator~\cite{he2024id} and ConsisID~\cite{yuan2024identity}, which exhibits limitations in motion variety and prompt adherence.

To complement our quantitative metrics, we conducted a comprehensive user study to evaluate the perceptual quality of generated results. The study involved 173 participants who assessed the outputs across four key aspects: motion dynamics, text-motion alignment, video quality, and identity consistency. Participants rated each aspect on a scale of 1-10, with results summarized in~\cref{tab:userstudy}. As shown in the overall preference scores in~\cref{tab:quantitative_result}, MagicMirror consistently outperforms baseline methods across most evaluated dimensions, demonstrating its superior perceptual quality in human assessment. Regarding video quality and ID similarity, we observed a gap between designed metrics and human-evaluated perceptual evaluation in ConsisID~\cite{yuan2024identity}.

\begin{table}[t]
	\centering
        \scriptsize
	\renewcommand\arraystretch{1}
	{\setlength{\tabcolsep}{2.3pt}
		\begin{tabular}{@{}l|ccccc|ccc@{}}
			\toprule
			Exp. & $\hat{x}_\text{id}$ & $x_\text{face}$ & $m_\text{face}$ & $\hat{m}$ & Pretrain & txt-align\rf{$^\uparrow$} & FM$_\text{inter}\rf{^\uparrow}$ & ID\rf{$^\uparrow$}\\
            \midrule
 			A [identity branch] & \ding{51} & & & & & 0.238 & 0.572 & 0.865\\
                B [structural branch] & & \ding{51} & & & & 0.240 & 0.584 & 0.869\\
                C [dual branch]& \ding{51} & \ding{51} & & &  & 0.239 & 0.654 & 0.870\\
                D [Eq.\ref{eq:ln}, $m_\text{face}$]&  & \ding{51} & \ding{51} & && 0.241 & 0.563 & 0.872\\
                E [Eq.\ref{eq:ln_add}, $\hat{m}_\text{txt}, \hat{m}_\text{vid}$]& \ding{51} & & & \ding{51} &  & 0.242 & 0.696 & 0.875\\
                F [w/o CAN]& \ding{51} & \ding{51} & & &\ding{51} & 0.236 & 0.568 & 0.886\\
                G [w/o pretrain]& \ding{51} & \ding{51} &\ding{51} &\ding{51} & & 0.241 & 0.559 & 0.883\\
                % G &  &  & \ding{51} & \ding{51} & \ding{51} \\
                \cellcolor{Gray}Full [MagicMirror] & \cellcolor{Gray}\ding{51} & \cellcolor{Gray}\ding{51} & \cellcolor{Gray}\ding{51} & \cellcolor{Gray}\ding{51} & \cellcolor{Gray}\ding{51} & \cellcolor{Gray}0.240 & \cellcolor{Gray}0.665 & \cellcolor{Gray}0.911\\
			\bottomrule
		\end{tabular}}
        \vspace{-5pt}
        \caption{\textbf{Ablation study results} on the same training scale across multiple settings. Some modules are interdependent.}
        \vspace{-15pt}
        \label{tab:ablation}
\end{table}
\vspace{-1mm}
\subsection{Ablation Studies}
\vspace{-1mm}
\paragraph{Condition-related Modules.}
% We evaluate our key architectural components through ablation studies, shown in ~\cref{fig:ablation}. Without the reference feature embedding branch, the model loses crucial high-level attention guidance, significantly degrading identity fidelity. The conditioned adaptive normalization~(CAN) proves vital for distribution alignment, enhancing identity preservation across frames. The effectiveness of CAN for facial condition injection is further demonstrated in~\cref{fig:convergence_can}, showing improved training convergence for identity information capture during the image pre-train stage.
We evaluate our key modules through ablation studies, shown in ~\cref{tab:ablation} and ~\cref{fig:ablation}. To ensure a fair comparison, all ablations are conducted on the same training scale, with a half training iteration of the official setting. Experiments using single-branch facial embedding (Exp. A: identity branch only, Exp. B: structural branch only) exhibit a limited identity preservation (0.865-0.869 ID similarity). In contrast, the dual-branch strategy (Exp. C) synergizes their complementary strengths, and achieves a higher motion metric.
The Conditioned Adaptive Normalization~(CAN) proves vital for distribution alignment, enhancing identity preservation across frames. The effectiveness of CAN for facial condition injection is further demonstrated in Exp. D, E, and F. Notably, complete CAN removal (Exp. F) causes significant performance degradation (0.911 $\rightarrow$ 0.886), underscoring its necessity for effective identity injection. An extended analysis of the CAN's benefits for the training convergence and distribution alignment is provided in the Appendix \textcolor{iccvblue}{B.1-B.2}.

\paragraph{Training Strategy.}
Exp. G in~\cref{tab:ablation} and \cref{fig:ablation} also illustrate the impact of different training strategies. Image pre-training is essential for robust identity preservation, while video post-training ensures temporal consistency. 
% However, training exclusively on image data leads to color-shift artifacts during video inference. This artifact is caused by modulation factor inconsistencies in different training stages. 
Our two-stage training approach achieves optimal results by leveraging the advantages of both phases, generating high ID fidelity videos with dynamic facial motions. 
Appendix \textcolor{iccvblue}{B.3} discusses more details about the training strategy.

\vspace{-1mm}
\section{Conclusion}
\vspace{-1mm}
\label{sec:conclusion}
In this work, we presented MagicMirror, a zero-shot framework for ID-preserving video generation. MagicMirror incorporates dual facial embeddings and Conditional Adaptive Normalization~(CAN) into DiT-based architectures. Our approach enables robust identity preservation and stable training convergence. Extensive experiments demonstrate that MagicMirror generates high-quality personalized videos while maintaining identity consistency from a single reference image, outperforming existing methods across multiple benchmarks and human evaluations. 

\paragraph{Limitations.}
While MagicMirror excels at ID-consistent video generation, challenges remain in supporting multiple identities and preserving fine-grained attributes beyond facial features, such as clothing, improvements necessary for practical customized video applications.

{
\paragraph{Acknowledgment.} The study was supported in part by the Research Grants Council under the Areas of Excellence scheme grant AoE/E-601/22-R, Hong Kong General Research Fund (14208023), Hong Kong AoE/P-404/18, and the
Center for Perceptual and Interactive Intelligence (CPII) Ltd under InnoHK supported by the Innovation and Technology Commission.
}

\begin{comment}{
\paragraph{Acknowlegdement} This work was supported in part by the Research Grants Council under the Areas of Excellence scheme grant AoE/E-601/22-R and the Shenzhen Science and Technology Program under No.~KQTD20210811090149095.
}
\end{comment}
% \newpage
% \input{sec/2_formatting}
% \input{sec/3_finalcopy}
%\small \bibliographystyle{ieeenat_fullname} %\bibliography{main}

%
\newpage
\clearpage
\appendix
\etoctoccontentsline{part}{Supplementary Material}
% \setcounter{page}{1}
% \maketitlesupplementary

\section*{Appendix}
This appendix provides comprehensive technical details and additional results for MagicMirror, encompassing dataset preparation, architectural specifications, implementation, and extensive experimental validations. We include additional qualitative results and in-depth analyses to support our main findings. \textbf{We strongly encourage readers to examine the project page}
{\small\textbf{\url{https://julianjuaner.github.io/projects/Magic-Mirror/}}}
for dynamic video demonstrations. The following contents are organized for efficient navigation.

\setlength{\cftbeforesecskip}{0.5em}
\cftsetindents{section}{0em}{1.8em}
\cftsetindents{subsection}{1em}{2.5em}
\cftsetindents{subsubsection}{3.0em}{3.5em}

\renewcommand{\contentsname}{Appendix Contents}
\hypersetup{linkbordercolor=black,linkcolor=black}
\localtableofcontents
\hypersetup{linkbordercolor=iccvblue,linkcolor=iccvblue}

\begin{figure*}[t]
    \centering
    \includegraphics[width=1\linewidth]{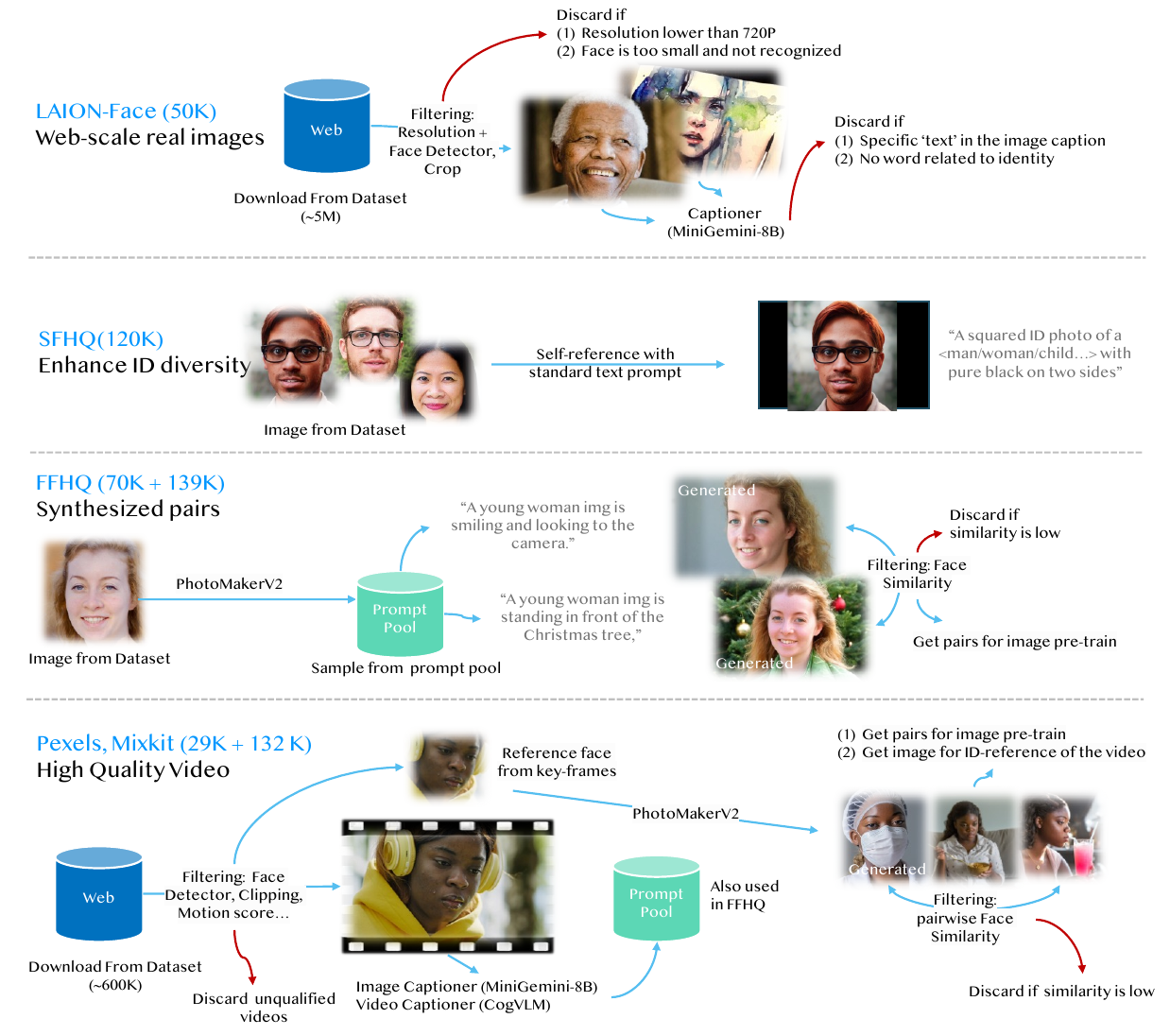}
    %\vspace{-2mm}
    \caption{\textbf{Detailed training data processing pipeline.} Building upon~\cref{fig:data}, we illustrate comprehensive filtering criteria, prompt examples, and processing specifications. The data flow is indicated by \textcolor{iccvblue}{blue arrows}, while filtering rules leading to data exclusion are marked with \textcolor{red}{red arrows}.}
    \label{fig:supp_data}
    %\vspace{-2mm}
\end{figure*}

%\vspace{-1mm}
\section{Experiment Details}\label{sec:exp_detail}
%\vspace{-1mm}
\subsection{Training Data Preparation}\label{sec:train_collection}
%\vspace{-1mm}
Our training dataset is constructed through a rigorous preprocessing pipeline, as illustrated in~\cref{fig:supp_data}. For the image pretrain data, we start downloading 5 million images from LAION-face~\cite{schuhmann2021laion}, then undergo strict quality filtering based on face detection confidence scores and resolution requirements. The filtered subset of 107K images is then processed through an image captioner~\cite{li2024mini}, where we exclude images containing texts. This results in a curated set of 50K high-quality face image-text pairs.
To enhance identity diversity, we incorporate the synthetic SFHQ dataset~\cite{david_beniaguev_2022_SFHQ}. To fit the model output, we standardize these images by adding black borders and pairing them with a consistent prompt template: \textit{"A squared ID photo of ..., with pure black on two sides."} This preprocessing ensures uniformity while maintaining the dataset's diverse identity characteristics.

For FFHQ~\cite{karras2019style}, we leverage a state-of-the-art identity-preserving prior PhotoMakerV2~\cite{li2024photomaker} to generate synthetic images of the same identity, but with different face poses. We filter redundant identities using pairwise facial similarity metrics, with prompts sampled from our 50K video keyframe captions.
We use the Pexels-400K and Mixkit datasets from~\cite{lin2024opensoraplanopensourcelarge} for construction of image-video pairs. The videos undergo a systematic preprocessing pipeline, including face detection and motion-based filtering to ensure high-quality dynamic content. We generate video descriptions using CogVLM video captioner~\cite{wang2023cogvlm}. Following our FFHQ processing strategy, we employ PhotoMakerV2 to synthesize identity-consistent images from the detected faces, followed by quality-based filtering.

%\vspace{-1mm}
\subsection{Test Data Preparation}\label{sec:testdata_collection}
%\vspace{-1mm}
\paragraph{Face Images Preparation} We construct a comprehensive evaluation set for identity preservation assessment across video generation models. Our dataset comprises 50 distinct identities across seven demographic categories: man, woman, elderly man, elderly woman, boy, girl, and baby. The majority of faces are sourced from PubFig dataset~\cite{kumar2009attribute}, supplemented with public domain images for younger categories. Each identity is represented by 1-4 reference images to capture variations in pose and expression.

\begin{figure*}[ht]
    \centering
    \includegraphics[width=1\linewidth]{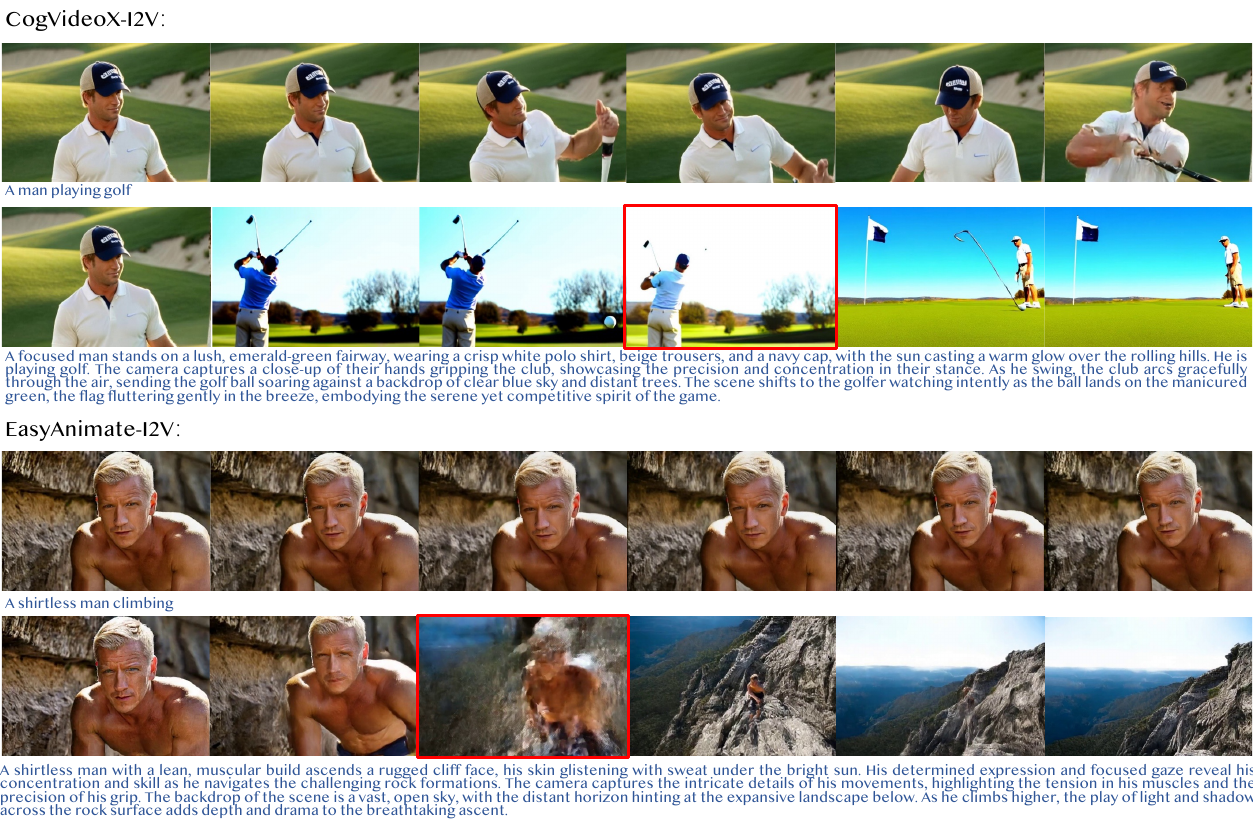}
    %\vspace{-5mm}
    \caption{\textbf{Impact of prompt length on image-to-video generation.} We demonstrate how image-to-video models perform differently with concise versus enhanced prompts. Frames with large artifacts are \textcolor{red}{marked in red}. First frame images are generated from enhanced prompts.}
    \label{fig:comparation}
    %\vspace{-2mm}
\end{figure*}

\paragraph{Prompt Preparation} Our test prompts are derived from VBench~\cite{huang2024vbench}, focusing on human-centric actions. For detailed descriptions, we sample from the initial 200 GPT-4-enhanced prompts and select 77 single-person scenarios. Each prompt is standardized with consistent subject descriptors and augmented with the \texttt{img} trigger word for model compatibility. We assign four category-appropriate prompts to each identity, ensuring demographic alignment. For the "baby" category, which lacks representation in VBench, we craft four custom prompts to maintain evaluation consistency across all categories.

%\vspace{-1mm}
\subsection{Comparisons}
%\vspace{-1mm}
\paragraph{ID-Animator~\cite{he2024id}} We utilize enhanced long prompts for evaluation, although some undergo partial truncation due to CLIP's 77-token input constraint.

In our main comparisons ~\cref{tab:quantitative_result}, we evaluated ID-Animator at a resolution of 480×720. This choice was made to ensure that SD-based ID-Animator comparisons used matching resolutions, thereby ensuring equal content capacity—a decision justified by the inherent resolution independence of the UNet architecture. To provide a fair and comprehensive evaluation, we additionally present some results at the default 512×512 resolution here in ~\cref{tab:id-animator}. These results confirm that our comparisons remain robust and consistent across different resolution settings.

\vspace{-9pt}
\begin{table}[H]
	\centering
        \scriptsize
	\renewcommand\arraystretch{1}
	{\setlength{\tabcolsep}{3pt}
		\begin{tabular}{@{}c|cc|cccc@{}}
			\toprule
			Method & Base & Resolution & txt-align\rf{$^\uparrow$} & FM$_\text{inter}\rf{^\uparrow}$ & ID\rf{$^\uparrow$} & Smooth\rf{$^\downarrow$}\\
            \midrule
 			ID-Animator & SD1.5 & (480, 720) & 0.211 & 0.181 & 0.923 & 0.515\\
                ID-Animator & SD1.5 & (512, 512) & 0.217 & 0.179 & 0.921 & 0.501\\
                % ConsisID    & CogVideoX & (480, 720) & 0.236  & 2.29 & 0.913 &0.513\\ 
                \cellcolor{Gray} MagicMirror & \cellcolor{Gray}CogVideoX & \cellcolor{Gray}(480, 720) &  \cellcolor{Gray}0.240 & \cellcolor{Gray}0.610 & \cellcolor{Gray}0.922 & \cellcolor{Gray}0.484\\
			\bottomrule
		\end{tabular}}
        \vspace{-5pt}
        \caption{\textbf{ID-Animator resolution comparison.}}
        \vspace{-11pt}
        \label{tab:id-animator}
\end{table}

\paragraph{ConsisID~\cite{yuan2024identity}} We utilize the CogVideoX-5B\cite{yang2024cogvideox} version. Its base inference settings are aligned with those of our model, and enhanced long prompts are employed to fully leverage its capabilities.

\paragraph{CogVideoX-5B-I2V~\cite{yang2024cogvideox}} For this image-to-video variant, we first generate reference images using PhotoMakerV2~\cite{li2024photomaker} for each prompt-identity pair. These images, combined with enhanced long prompts, serve as input for video generation.

\paragraph{EasyAnimate~\cite{xu2024easyanimate}} We evaluate using the same PhotoMakerV2-generated reference images as in our CogVideoX-5B-I2V experiments.

\paragraph{DynamiCrafter~\cite{xing2023dynamicrafter}} Due to model-specific resolution requirements, we create a dedicated set of reference images using PhotoMakerV2 that conform to the model's specifications.

In image-to-video baselines, through reference images generated by enhanced prompts, we deliberately use original short concise prompts for video generation. This choice stems from our empirical observation that image-to-video models exhibit a strong semantic bias when processing lengthy prompts. Specifically, these models tend to prioritize text alignment over reference image fidelity, leading to degraded video quality and compromised identity preservation. This trade-off is particularly problematic for our face preservation objectives. We provide visual evidence of this phenomenon in~\cref{fig:comparation}.

%\vspace{-1mm}
\subsection{Evaluation Metrics}
%\vspace{-1mm}
Our evaluation framework combines standard video metrics with face-specific measurements. From VBench~\cite{huang2024vbench}, we utilize Dynamic Degree for motion naturality and Overall Consistency for text-video alignment. Video quality is assessed using Inception Score from EvalCrafter~\cite{liu2024evalcrafter}. For facial fidelity, we measure identity preservation using facial recognition embedding similarity~\cite{geitgey2017facerecognition} and temporal stability through frame-wise similarity decay.

\begin{figure}
    \centering
    \includegraphics[width=1\linewidth]{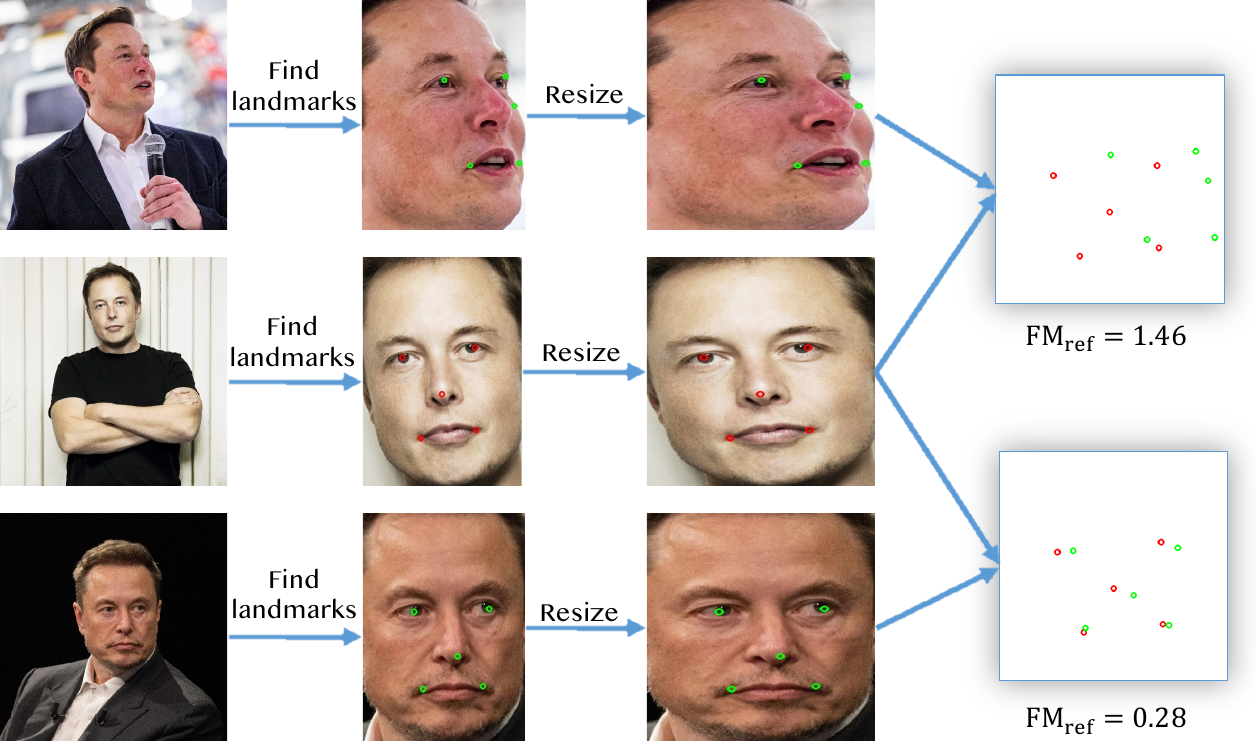}
    %\vspace{-4mm}
    \caption{\textbf{Face Motion (FM) calculation.} FM$_\text{inter}$ follows a similar computation across consecutive video frames.}
    \label{fig:movement}
    %\vspace{-4mm}
\end{figure}

We propose a novel facial dynamics metric to address the limitation of static face generation in existing methods. As shown in \cref{fig:movement}, we extract five facial landmarks using RetinaFace~\cite{deng2020retinaface} and compute two motion scores: FM$_\text{ref}$ measures facial motion relative to the reference image (computed on aspect-ratio-normalized frames to eliminate positional bias), while FM$_\text{inter}$ quantifies maximized inter-frame facial motion (computed on original frames to preserve translational movements). This dual-score approach enables a comprehensive assessment of facial dynamics.

\paragraph{Success rate \& failcase analysis.} Success rate metrics better demonstrate reliability. Our additional experiments with 200 videos (50 prompts × 4 seeds) compared MagicMirror with identity preservation baselines on success rates for face recognition, identity check, motion, and text alignment. MagicMirror achieves improved SR across most dimensions, though failcase analysis reveals motion quality remains the primary limitation, which is predominantly model-dependent. 
%The results confirm our method's reliability and will be included in the final paper. 

\begin{table}[H]
	\centering
        \resizebox{1.0\linewidth}{!}{
	\renewcommand\arraystretch{1}
	{\setlength{\tabcolsep}{2.7pt}
		\begin{tabular}{@{}l|cccc|c@{}}
			\toprule
			Method / SR & motion quality &  text align & face recongized & identity check & average\\
            \midrule
                ID-Animator & 11.5\% & 60.0\% & 93.5\% & 82.5\% & 61.9\% \\
                ConsisID  & 38.5\% & 73.5\%  & 89.5\% & 74.5\% & 69.0\%\\ 
                \cellcolor{Gray}MagicMirror & \cellcolor{Gray}44.0\% & \cellcolor{Gray}75.5\% &  \cellcolor{Gray}98.0\% & \cellcolor{Gray}81.0\% & \cellcolor{Gray}\textbf{74.6\%}\\
			\bottomrule
		\end{tabular}}
        }
        \label{loss_table}
        \caption{\textbf{Success rate comparison.}}
\end{table}

%\vspace{-1mm}
\subsection{Implementation Details}
%\vspace{-1mm}
\paragraph{Decoupled Facial Embeddings.} Our architecture employs two complementary branches: an ID embedding branch based on pre-trained PhotoMakerV2~\cite{li2024photomaker} with two-token ID-embedding query $q_\text{id}$, and a facial structural embedding branch that extracts detailed features from the same ViT's penultimate layer. The latter initializes 32 token embeddings as facial query $q_\text{face}$ input. We use a projection layer to align facial modalities before diffusion model input.

\begin{figure}
    \centering
    \includegraphics[width=1\linewidth]{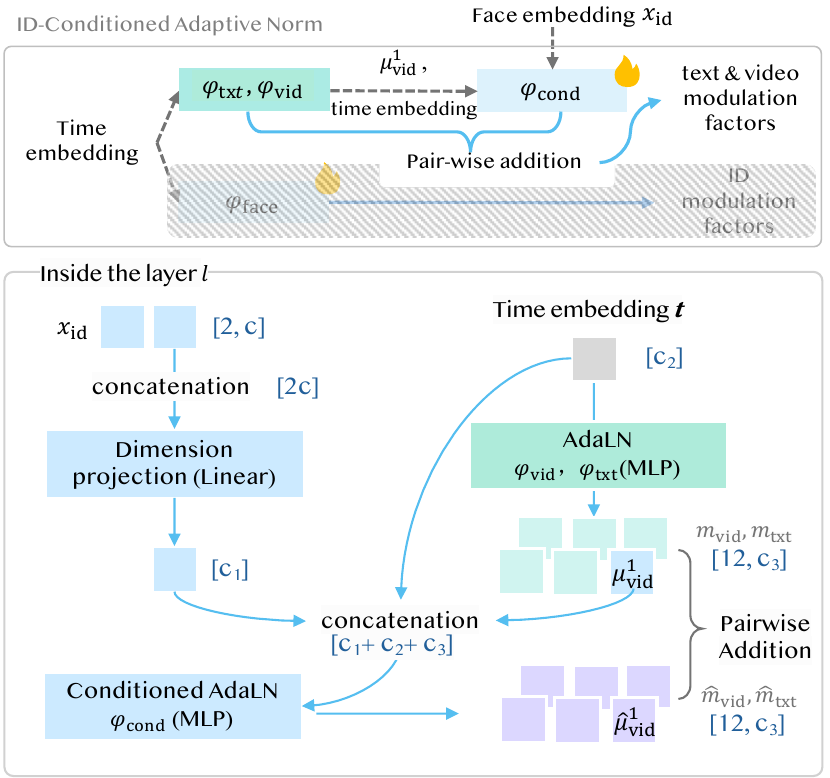}
    %\vspace{-4mm}
    \caption{\textbf{Detailed implementation of Conditioned Adaptive Normalization.} We present the expanded architecture of $\varphi_\text{cond}$~(illustrated in the unmasked region above) with comprehensive annotations of input-output tensor dimensions at each transformation.}
    \label{fig:supp_CAN}
    %\vspace{-4mm}
\end{figure}

\paragraph{Conditioned Adaptive Normalization.}
This paragraph elaborates on the design details of the Conditioned Adaptive Normalization~(CAN) module, complementing the overview provided in~\cref{sec:CAN} and~\cref{fig:adapter}. For predicting facial modulation factors $m_\text{face}$, we employ a two-layer MLP architecture, following the implementation structure of the original normalization modules $\varphi_\text{\{vid, text\}}$. The detailed implementation of CAN is illustrated in~\cref{fig:supp_CAN}. Given the facial ID embedding $x_\text{id}\in \mathcal{R}^{2\times c}$ containing two tokens, we first apply one global projection layer for dimensionality reduction, mapping it to dimension $c_1$. Subsequently, in each adapted layer, we concatenate this projected embedding with the time embedding $\mathbf{t}$ and the predicted shift factor $\mu^{1}_\text{vid}$ along the channel dimension. An MLP then processes this concatenated representation to produce the final modulation factors. To ensure stable training, all newly introduced modulation predictors are initialized with zero. 

We also tried to directly use the prediction of CAN as the data distribution, this results in a bad initialization, comparing with the residual prediction, direct prediction leads to abnormal video generation quality.

%\vspace{-1mm}
\section{Additional Discussions}
%\vspace{-1mm}
\subsection{Advantages of CAN}
\begin{figure}[t]
    \centering
    \includegraphics[width=1.0\linewidth]{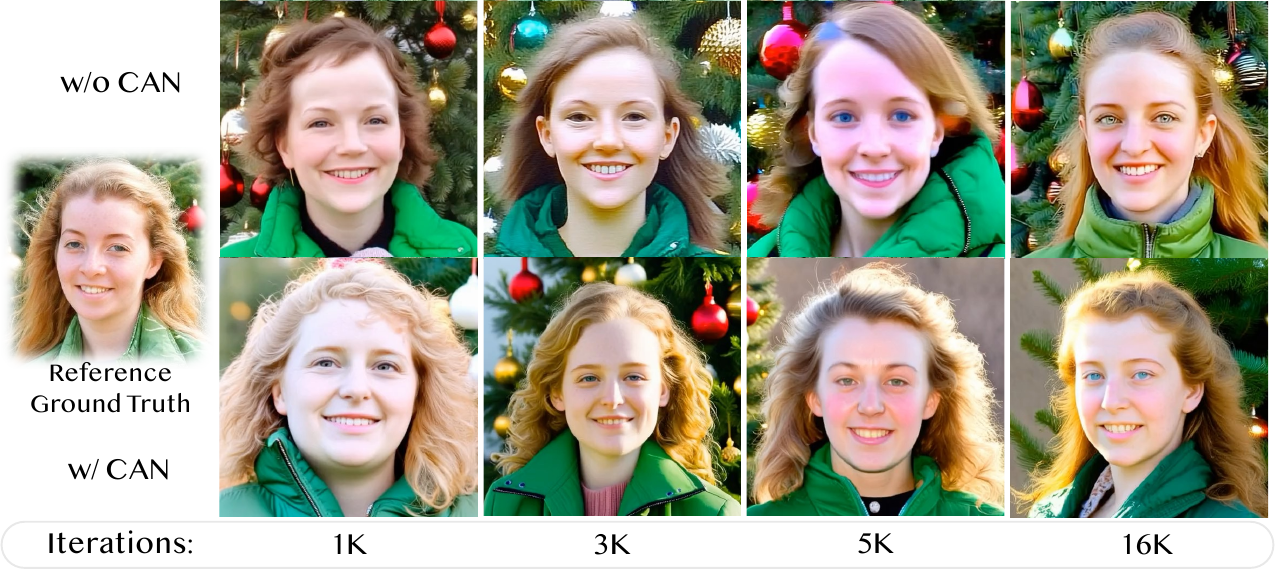}
    %\vspace{-2mm}
    \caption{\textbf{CAN speeds up the convergence.} Without the Conditioned Adaptive Normalization, the model cannot fit the simplest appearance features like hairstyle in the image pre-train stage.}
    \label{fig:convergence_can}
    % \vspace{-3mm}
\end{figure}

The benefits of CAN in facial condition injection are evident in its ability to enhance training convergence, particularly during the image pre-train stage. As illustrated in~\cref{fig:convergence_can}, models equipped with CAN achieve significantly improved identity information capture, enabling faster adaptation to appearance attributes. This acceleration in convergence highlights CAN's effectiveness in preserving identity consistency throughout the training process.

Furthermore, we specifically design CAN and related modules to be lightweight and avoid altering \textbf{any} pre-trained weights of the video DiT, thereby preserving the original model capacity. We evaluate GPU memory utilization, parameter count, and inference latency for generating a 49-frame 480P video. Compared to the baseline model, the additional parameters introduced by MagicMirror are primarily concentrated in the embedding extraction stage, which requires only a single forward pass. As summarized in~\cref{tab:supp_computation}, compared with ConsisID~\cite{yuan2024identity} and CogVideoX~\cite{yang2024cogvideox} baseline, MagicMirror introduces minimal computational overhead, with only a slight increase in GPU memory consumption and inference time. 

\begin{table}[H]
	\centering
        \resizebox{1\linewidth}{!}{
	\renewcommand\arraystretch{1.03}
	{\setlength{\tabcolsep}{2pt}
	\begin{tabular}{@{}l|cccccccc@{}}
	\toprule
        Model& Video size & Memory & Params. & Latency & Batch$\times$Iter. & Data (I+V) & GPU \\
        \midrule
        ID-Animator & (16,512,512) & 8.4~GiB & 1.52B & 11s & 2$\times$58K & 0K+13K & A100*1 \\
        \midrule
        CogVideoX-5B & (49,480,720) &24.9~GiB & 10.5B & 204s & (0.1-2K)$\times$750K & 2B+35M & - \\
        \midrule
        CogVideoX-I2V & (49,480,720) &25.9~GiB & 10.6B & 213s & - & - & -\\
        ConsisID & (49,480,720) &41.5~GiB & 11.1B & 213s& 80$\times$1.8K & 0K+130K & H100*40\\
        \cellcolor{Gray}MagicMirror  & \cellcolor{Gray}(49,480,720) & \cellcolor{Gray}28.6~GiB & \cellcolor{Gray}12.8B & \cellcolor{Gray}209s & \cellcolor{Gray}8$\times$9K* & \cellcolor{Gray}570K +29K & \cellcolor{Gray}A800*8\\
        \bottomrule
	\end{tabular}}
        }
        \caption{\textbf{Computation overhead of MagicMirror.} All computations are measured on the one A800 GPU.}
    \label{tab:supp_computation}
\end{table}

\begin{comment}
\begin{table}[]
{
    \tablestyle{5pt}{1.1}
    \centering
    \begin{tabular}{y{77}|x{40}x{40}x{40}}
        \toprule
        Model& Memory & Parameters & Time \\
        \midrule
        CogVideoX-5B~\cite{yang2024cogvideox} & 24.9~GiB & 10.5B & 204s\\
        \midrule
        ConsisID~\cite{yuan2024identity} & 41.5~GiB & 11.1B & 213s\\
        \cellcolor{Gray}MagicMirror  & \cellcolor{Gray}28.6~GiB & \cellcolor{Gray}12.8B & \cellcolor{Gray}209s \\
        \bottomrule
    \end{tabular}
    \caption{\textbf{Computation overhead of MagicMirror.} All computations are measured on the one A800 GPU.}
    \label{tab:supp_computation}
}
\end{table}
\end{comment}

%\vspace{-1mm}
\subsection{Distribution Analysis and Its Impact}
%\vspace{-1mm}
\begin{figure}[t]
    \centering
    \includegraphics[width=0.92\linewidth]{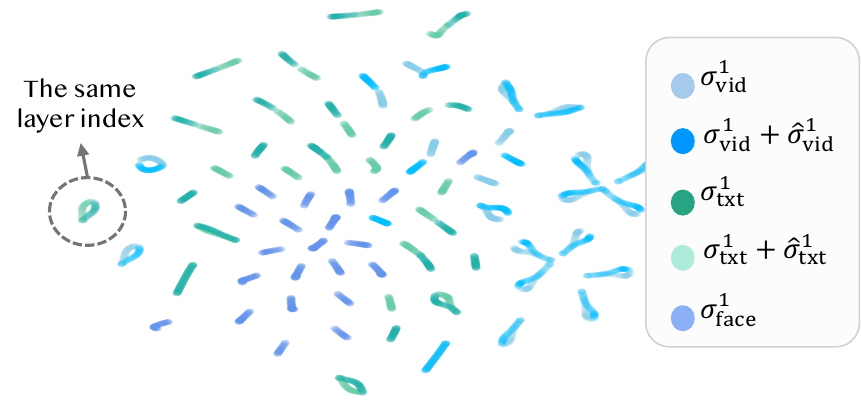}
    %\vspace{-1mm}
    \caption{
    %\textit{Left}: ID-Reference similarity curve among generated frame indexes for I2V and training-based methods. 
    \textbf{Different modalities' scale distribution using t-SNE.} Each point represents the scale with a unique timestep-layer index. We also illustrate a shift variant on text and video's adaptive scale using different colors.}
    \label{fig:similarity}
    %\vspace{-2mm}
\end{figure}

\begin{figure}
    \centering
    \includegraphics[width=1\linewidth]{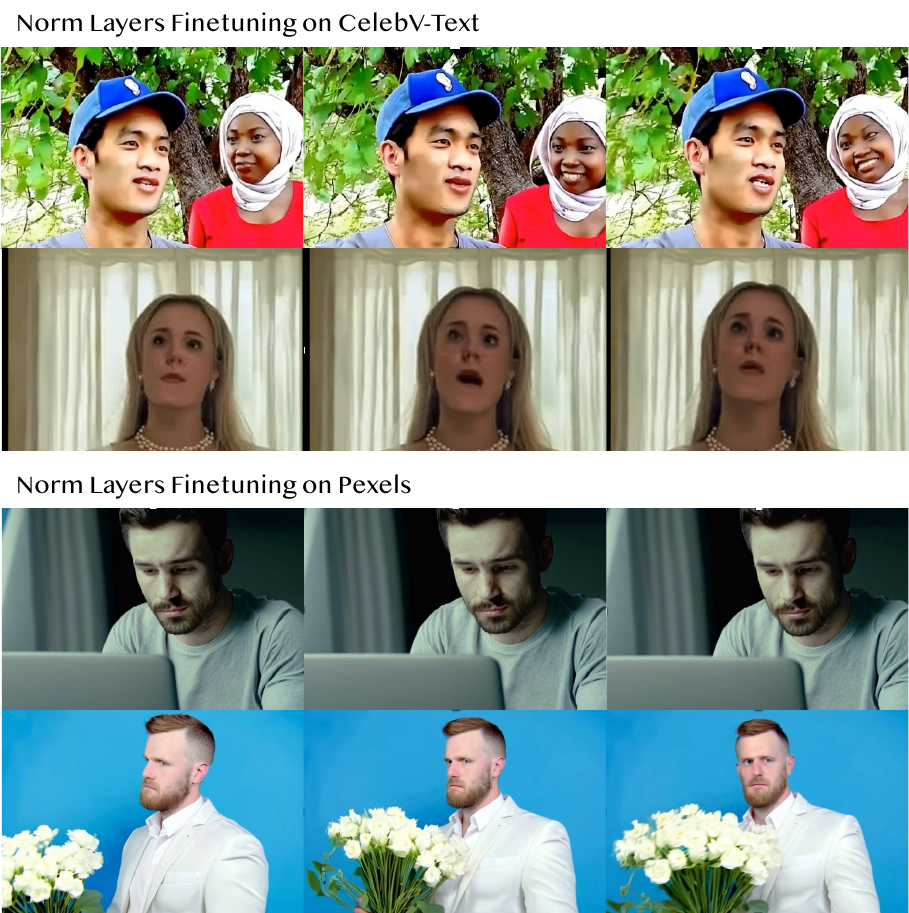}
    %\vspace{-4mm}
    \caption{\textbf{Modulation layers reflect data distribution.} Fine-tuning solely the modulation layer weights demonstrates adaptation to distinct data distributions, affecting both spatial fidelity and temporal dynamics.}
    \label{fig:supp_quality}
    %\vspace{-4mm}
\end{figure}

We begin by visualizing the predicted modulation scale factors \(\sigma\) using t-SNE~\cite{van2008visualizing} in \cref{fig:similarity}. The results show that distinct modalities occupy characteristic distributions across different Transformer layers, and these distributions appear largely invariant to the specific timestep input. In particular, the face modality exhibits a unique pattern, while the conditioned residual \(\hat{\sigma}\) introduces targeted shifts away from the baseline distribution. This shift empirically accelerates model convergence when incorporating ID conditions.

Beyond the t-SNE visualization, we further investigate the critical role of distribution alignment by examining how modality-aware data distributions affect generation quality. Specifically, we fine-tuned only the normalization layers \(\varphi_{\text{vid}}, \varphi_{\text{txt}}\) of the CogVideoX base model on two distinct datasets—CelebV-Text~\cite{yu2023celebv} and our Pexels video collection~\cite{pexels}. As illustrated in \cref{fig:supp_quality}, this distribution-specific fine-tuning exerts a substantial influence on the spatial fidelity of generated videos. 
These observations underscore the importance of aligning modality distributions during training, and they also validate the high quality of our curated video dataset.

Additionally, we conducted another experiment using our Pexels dataset. We found that by using a dataset with twice the frame rate and training only the modulation layers, we achieved an improvement in the VBench~\cite{huang2024vbench} dynamic motion score from 0.71 to 0.84. This result, similar to Experiments E-F in~\cref{tab:ablation}, further verifies the impact of the modulation module on dynamic facial motion.

\subsection{Two-Stage Training Analysis}
\begin{figure}[t]
    \centering
    \includegraphics[width=1.0\linewidth]{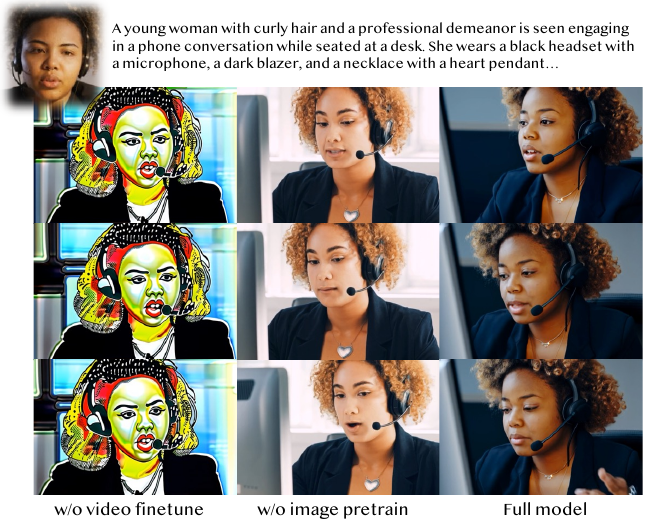}
    \vspace{-2mm}
    \caption{\textbf{Examples for ablation studies on training strategies.}}
    \label{fig:train_ablation}
    \vspace{-2mm}
\end{figure}

In \cref{fig:train_ablation}, we present additional ablation results that clarify how each training phase addresses a distinct aspect of identity-preserving video generation. Specifically, the image pre-training phase prioritizes robust identity encoding, ensuring that facial features remain consistent and accurately captured. However, training exclusively on image data leads to color-shift artifacts during video inference, caused by modulation factor inconsistencies across different training stages. By combining these two stages, our final approach aligns both identity representation and color distribution, resulting in dynamic and high-fidelity ID-preserving videos without the artifacts observed in single-stage alternatives.

% \subsection{Spatial Quality Gap}
% There are two 
%\vspace{-1mm}
\subsection{Limitation Analysis}
\begin{figure}[t]
    \centering
    \includegraphics[width=1\linewidth]{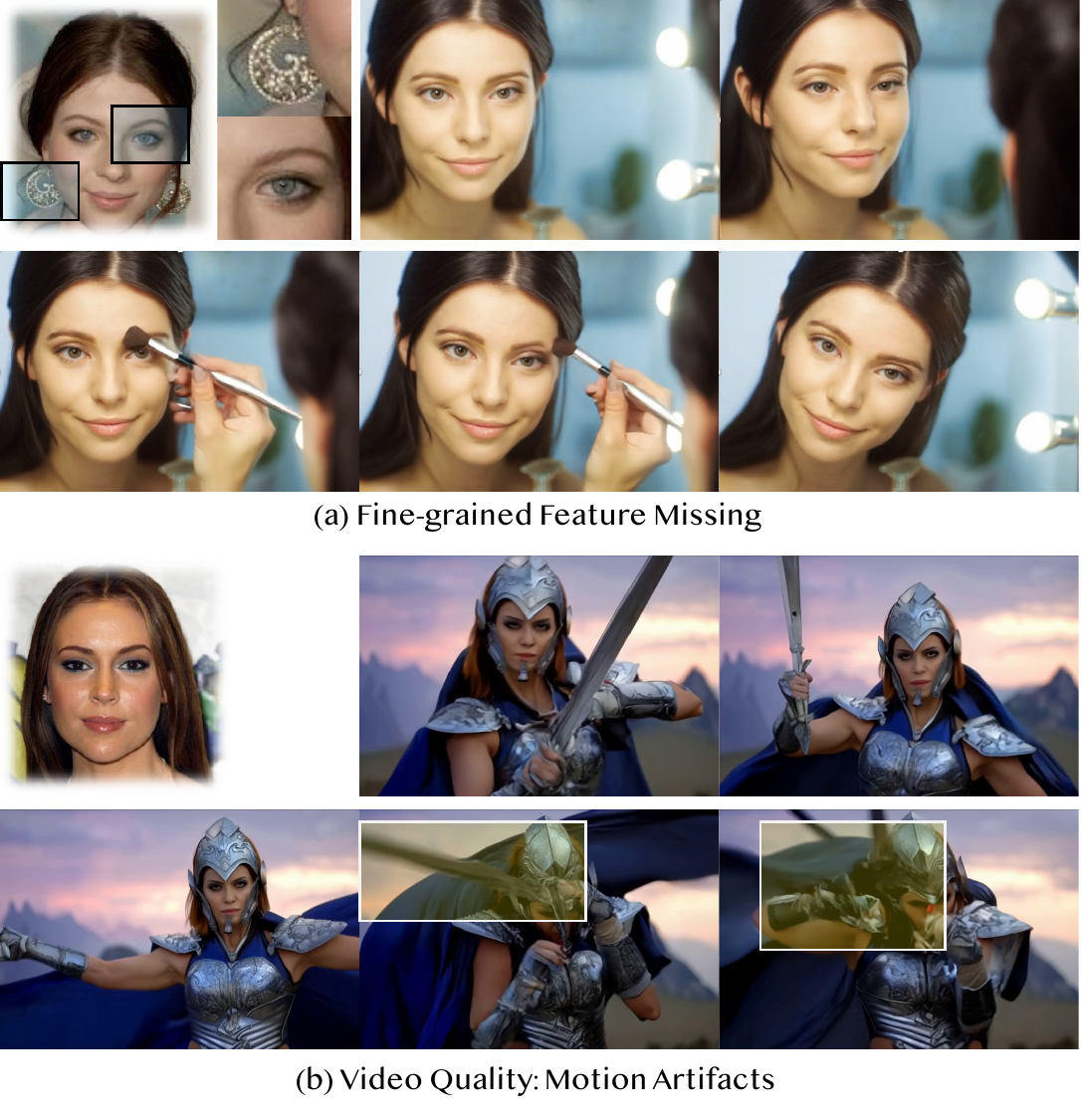}
    %\vspace{-4mm}
    \caption{\textbf{Limitations of MagicMirror.} (a) Fine-grained feature preservation failure in facial details and accessories. (b) Motion artifacts in generated videos showing temporal inconsistencies.}
    \label{fig:failcase}
%\vspace{-4mm}
\end{figure}

%\vspace{-1mm}
As discussed in~\cref{sec:conclusion}, our approach faces several limitations, particularly in handling multi-person scenarios and preserving fine-grained features. \cref{fig:failcase} illustrates two representative failure cases: incomplete transfer of reference character details (such as accessories) and motion artifacts caused by the base model. These limitations highlight critical areas for future research in controllable personalized video generation, particularly in maintaining temporal consistency and fine detail preservation.

%\vspace{-1mm}
\section{Additional Results \& Applications}
%\vspace{-1mm}
\subsection{Additional Applications}
\begin{figure}[ht]
    \centering
    \includegraphics[width=1\linewidth]{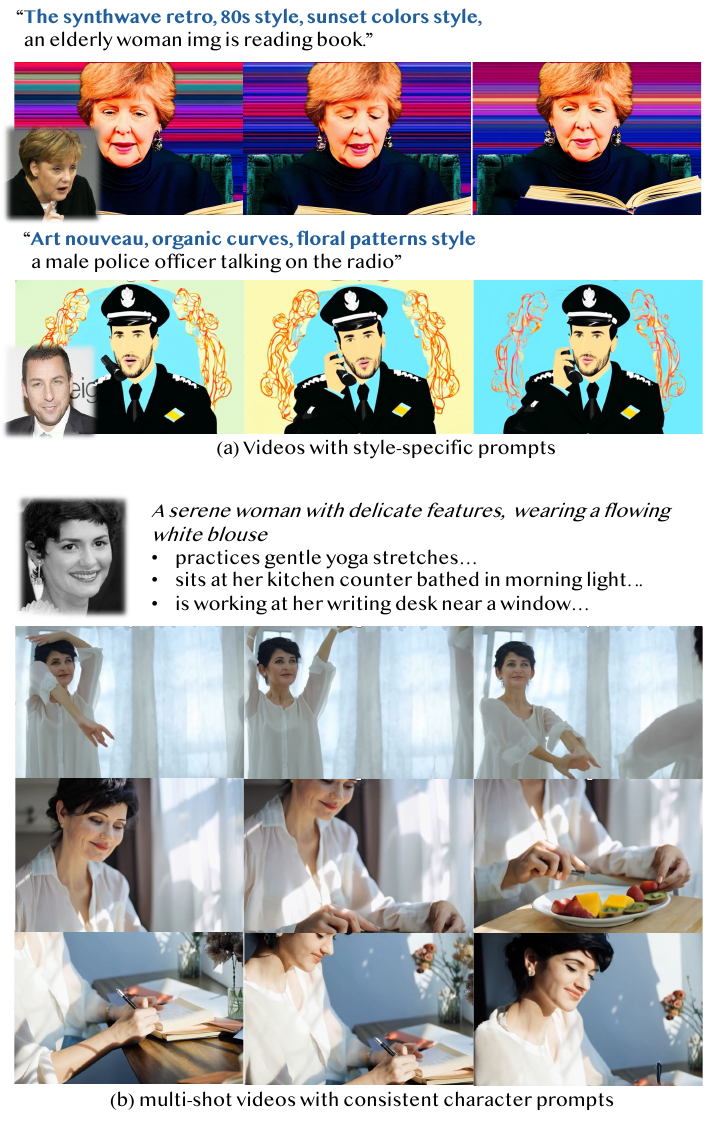}
    %\vspace{-4mm}
    \caption{\textbf{Additional applications of MagicMirror.} We can generate identity-preserved videos across artistic styles and can generate multi-shot videos with consistent characters. \textcolor{iccvblue}{More results are presented in the project page.}}
    \label{fig:supp_applications}
    %\vspace{-4mm}
\end{figure}

%\vspace{-1mm}
\cref{fig:supp_applications} demonstrates two extended capabilities of MagicMirror. First, beyond realistic customized video generation, our framework effectively handles stylized prompts, leveraging CogVideoX's diverse generative capabilities to produce identity-preserved outputs across various artistic styles and visual effects. Furthermore, we show that our method can generate high-quality, temporally consistent multi-shot sequences when maintaining coherent character and style descriptions. We believe these capabilities have significant implications for automated video content creation.

%\vspace{-1mm}
\subsection{Image Generation Results}
\begin{figure*}[p]
    \centering
    \includegraphics[width=0.99\linewidth]{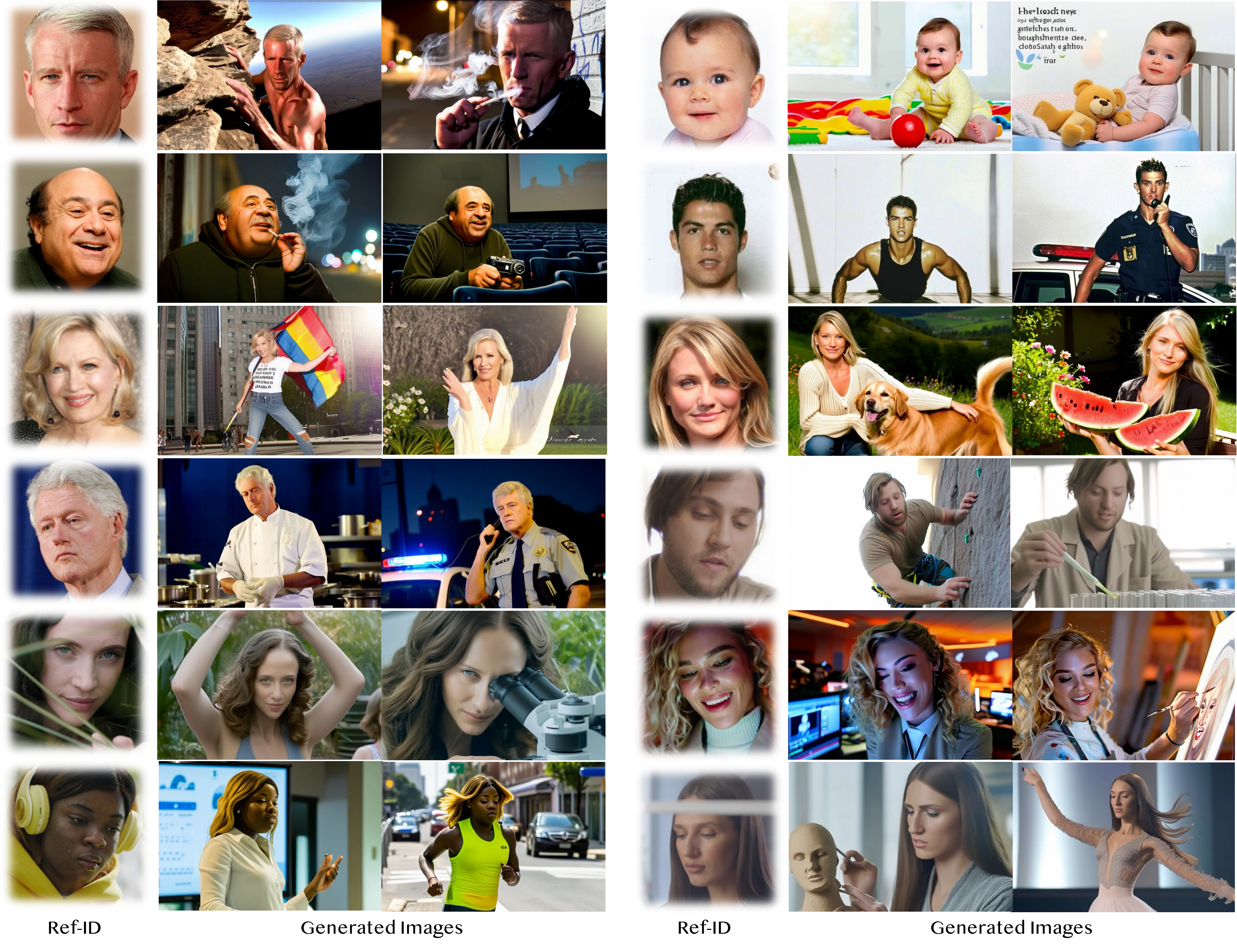}
    %\vspace{-2mm}
    \caption{\textbf{Image generation using MagicMirror.} Model in the image pre-train stage captures ID embeddings of the reference ID (Ref-ID), yet over-fits on some low-level distributions such as image quality, style, and background.}
    \label{fig:supp_img_gen}
    %\vspace{-4mm}
\end{figure*}

%\vspace{-1mm}
MagicMirror demonstrates strong capability in ID-preserving image generation with the image-pre-trained stage. Notably, it achieves even superior facial identity fidelity compared to video-finetuned variants, primarily due to optimization constraints in video training (e.g., limited batch sizes and dataset scope). Representative examples are presented in~\cref{fig:supp_img_gen}.

%\vspace{-1mm}
\subsection{Video Generation Results}
\begin{figure*}[p]
    \centering
    \includegraphics[width=0.99\linewidth]{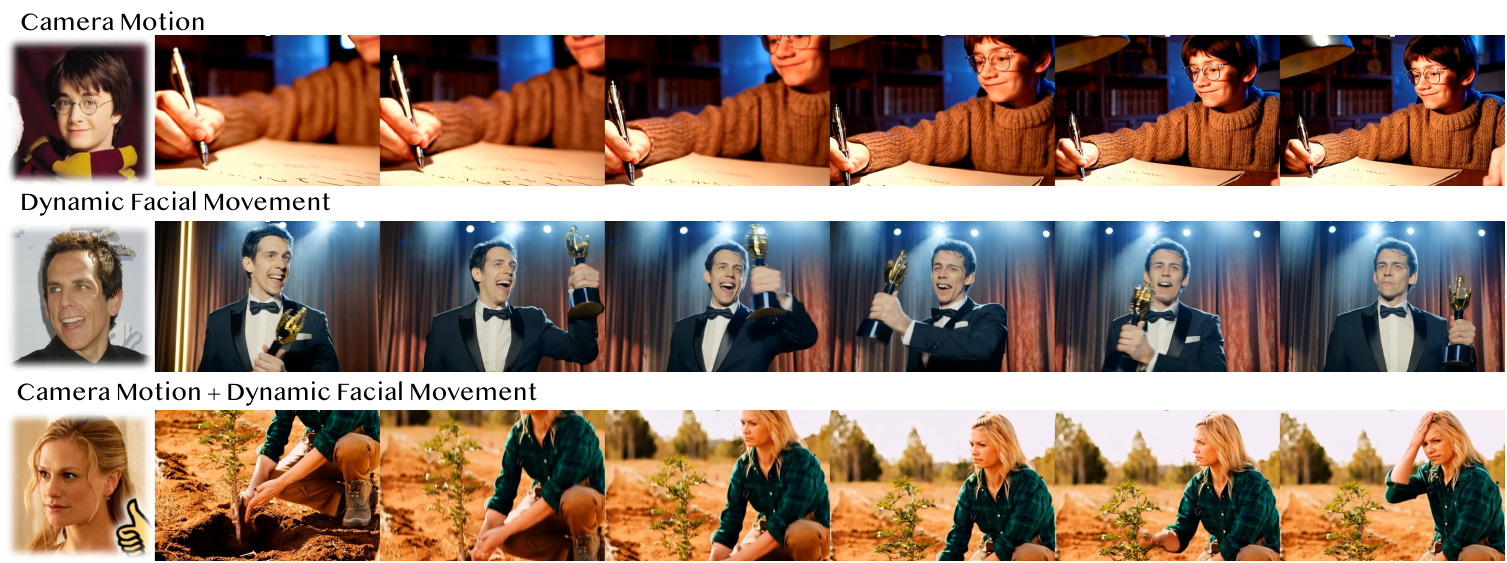}
    %\vspace{-2mm}
    \caption{\textbf{Advantages over I2V generation.} MagicMirror successfully handles challenging scenarios including partially occluded initial frames and maintains identity consistency through complex facial dynamics, addressing limitations of traditional I2V approaches.}
    \label{fig:supp_i2v_cant}
    %%\vspace{-4mm}
\end{figure*}

\begin{figure*}[ht]
    \centering
    \includegraphics[width=0.99\linewidth]{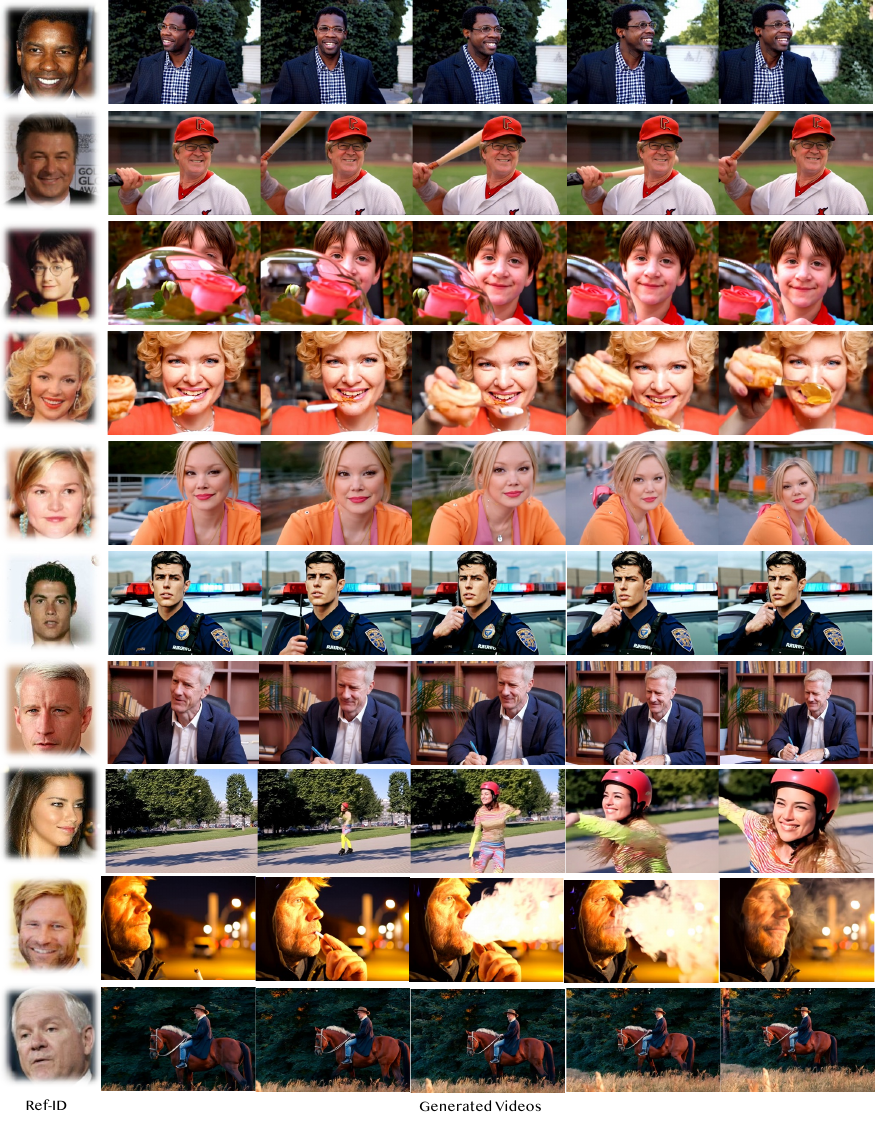}
    %\vspace{-2mm}
    \caption{\textbf{Video generation results.} We demonstrate MagicMirror's capability across varying facial scales and compositions. \textcolor{iccvblue}{Additional examples and comparative analyses are available in the project page.}}
    %\vspace{-2mm}
    \label{fig:supp_vid}
\end{figure*}

%\vspace{-1mm}
Additional video generation results and comparative analyses are provided in~\cref{fig:supp_i2v_cant,fig:supp_vid}, highlighting our method's advantages. \cref{fig:supp_i2v_cant} specifically demonstrates the benefits of our one-stage approach over I2V, including superior handling of occluded initial frames, enhanced dynamic range, and improved temporal consistency during facial rotations. In~\cref{fig:supp_vid}, we provide more results with human faces on different scales. 
% \textit{\textbf{More video results are available in the supplementary HTML/index.html.}}

%\vspace{-1mm}
\section{Acknowledgments}
%\vspace{-1mm}
\paragraph{Social Impact.} MagicMirror is designed to facilitate creative and research-oriented video content generation while preserving individual identities. We advocate for responsible use in media production and scientific research, explicitly discouraging the creation of misleading content or violation of portrait rights. As our framework builds upon the DiT foundation model, existing diffusion-based AI-content detection methods remain applicable.

\paragraph{Data Usage.} The training data we used is almost entirely sourced from known public datasets, including all image data and most video data. All video data was downloaded and processed through proper channels (i.e., download requests). We implement strict NSFW filtering during the training process to ensure content appropriateness.

\noindent{\textcolor{iccvblue}{\textit{\textbf{Figures 18-20 are presented on the following pages~$\downarrow$}}}}

\cleardoublepage

\small \bibliographystyle{ieeenat_fullname} \bibliography{main}
\end{document}